\documentclass[10pt,twocolumn,letterpaper]{article}

\usepackage{wacv}              

\usepackage{graphicx}
\usepackage{amsmath}
\usepackage{amssymb}
\usepackage{booktabs}
\usepackage{tabularray}
\usepackage{xcolor}
\usepackage{multirow}
\usepackage{multicol}
\usepackage{tabularx} 
\usepackage{pifont} 
\usepackage{verbatim} 
\usepackage{float} 
\usepackage[export]{adjustbox} 
\usepackage{siunitx}  
\usepackage{algorithmicx} 
\usepackage{algpseudocode} 
\usepackage{algorithm} 
\usepackage{subcaption}
\usepackage{xcolor}
\usepackage{overpic}
\definecolor{wacvblue}{rgb}{0.21,0.49,0.74}

\usepackage[pagebackref,breaklinks,colorlinks,allcolors=wacvblue]{hyperref}

\usepackage{tikz}
\usepackage{pgfplots}

\pgfplotsset{compat=newest}

\usetikzlibrary{arrows,arrows.meta, automata}
\usetikzlibrary{shapes.geometric,shapes.arrows,decorations.pathmorphing}
\usetikzlibrary{matrix,chains,scopes,positioning,arrows,fit}
\usetikzlibrary{patterns}
\usetikzlibrary{chains,backgrounds,calc}

\usetikzlibrary{shapes,arrows} 
\usetikzlibrary{calc} 

\definecolor{tu0}{rgb}{0.7451, 0.1176, 0.2353}

\definecolor{tu1}{rgb}{1.0000, 0.8039, 0.0000}
\definecolor{tu11}{rgb}{1.0000, 0.8627, 0.3020}
\definecolor{tu12}{rgb}{1.0000, 0.9020, 0.4980}
\definecolor{tu13}{rgb}{1.0000, 0.9412, 0.6980}
\definecolor{tu14}{rgb}{1.0000, 0.9608, 0.8000}

\definecolor{tu2}{rgb}{0.9804, 0.4314, 0.0000}
\definecolor{tu21}{rgb}{0.9882, 0.6039, 0.3020}
\definecolor{tu22}{rgb}{0.9882, 0.7137, 0.4980}
\definecolor{tu23}{rgb}{0.9922, 0.8275, 0.6980}
\definecolor{tu24}{rgb}{0.9961, 0.8863, 0.8000}

\definecolor{tu3}{rgb}{0.6902, 0.0000, 0.2745}
\definecolor{tu31}{rgb}{0.7529, 0.2000, 0.4196}
\definecolor{tu32}{rgb}{0.8431, 0.4980, 0.6353}
\definecolor{tu33}{rgb}{0.9216, 0.7490, 0.8196}
\definecolor{tu34}{rgb}{0.9529, 0.8510, 0.8902}

\definecolor{tu4}{rgb}{0.4863, 0.8039, 0.9020}
\definecolor{tu41}{rgb}{0.6431, 0.8627, 0.9333}
\definecolor{tu42}{rgb}{0.7412, 0.9020, 0.9490}
\definecolor{tu43}{rgb}{0.8431, 0.9412, 0.9686}
\definecolor{tu44}{rgb}{0.8980, 0.9608, 0.9804}

\definecolor{tu5}{rgb}{0.0000, 0.5020, 0.7059}
\definecolor{tu51}{rgb}{0.3020, 0.6510, 0.7961}
\definecolor{tu52}{rgb}{0.5490, 0.7765, 0.8667}
\definecolor{tu53}{rgb}{0.7490, 0.8745, 0.9255}
\definecolor{tu54}{rgb}{0.8510, 0.9255, 0.9569}

\definecolor{tu6}{rgb}{0.0000, 0.3255, 0.4549}
\definecolor{tu61}{rgb}{0.2510, 0.4941, 0.5922}
\definecolor{tu62}{rgb}{0.5490, 0.6941, 0.7529}
\definecolor{tu63}{rgb}{0.7490, 0.8314, 0.8627}
\definecolor{tu64}{rgb}{0.8510, 0.8980, 0.9176}

\definecolor{tu7}{rgb}{0.7765, 0.9333, 0.0000}
\definecolor{tu71}{rgb}{0.8431, 0.9529, 0.3020}
\definecolor{tu72}{rgb}{0.8863, 0.9647, 0.4980}
\definecolor{tu73}{rgb}{0.9333, 0.9804, 0.6980}
\definecolor{tu74}{rgb}{0.9569, 0.9882, 0.8000}

\definecolor{tu8}{rgb}{0.5373, 0.6431, 0.0000}
\definecolor{tu81}{rgb}{0.6784, 0.7490, 0.3020}
\definecolor{tu82}{rgb}{0.7686, 0.8196, 0.4980}
\definecolor{tu83}{rgb}{0.8588, 0.8941, 0.6980}
\definecolor{tu84}{rgb}{0.9059, 0.9294, 0.8000}

\definecolor{tu9}{rgb}{0.0000, 0.4431, 0.3373}
\definecolor{tu91}{rgb}{0.3020, 0.6118, 0.5373}
\definecolor{tu92}{rgb}{0.5490, 0.7490, 0.7020}
\definecolor{tu93}{rgb}{0.7490, 0.8588, 0.8353}
\definecolor{tu94}{rgb}{0.8549, 0.9176, 0.9059}

\definecolor{tu10}{rgb}{0.8000, 0.0000, 0.6000}
\definecolor{tu101}{rgb}{0.8706, 0.3490, 0.7412}
\definecolor{tu102}{rgb}{0.9216, 0.6000, 0.8392}
\definecolor{tu103}{rgb}{0.9608, 0.8000, 0.9216}
\definecolor{tu104}{rgb}{0.9804, 0.8980, 0.9608}

\definecolor{tu110}{rgb}{0.4627, 0.0000, 0.4627}
\definecolor{tu111}{rgb}{0.5961, 0.2510, 0.5961}
\definecolor{tu112}{rgb}{0.7294, 0.4980, 0.7294}
\definecolor{tu113}{rgb}{0.8392, 0.6980, 0.8392}
\definecolor{tu114}{rgb}{0.9216, 0.8510, 0.9216}

\definecolor{tu120}{rgb}{0.4627, 0.0000, 0.3294}
\definecolor{tu121}{rgb}{0.6118, 0.3020, 0.5333}
\definecolor{tu122}{rgb}{0.7569, 0.5490, 0.6980}
\definecolor{tu123}{rgb}{0.8667, 0.7490, 0.8314}
\definecolor{tu124}{rgb}{0.9216, 0.8510, 0.9020}

\definecolor{tu130}{rgb}{0.0314, 0.0314, 0.0314}
\definecolor{tu131}{rgb}{0.3725, 0.3725, 0.3725}
\definecolor{tu132}{rgb}{0.5882, 0.5882, 0.5882}
\definecolor{tu133}{rgb}{0.7529, 0.7529, 0.7529}
\definecolor{tu134}{rgb}{0.8667, 0.8667, 0.8667}
\newcommand{\VEC}[1]{\mathbf{#1}}          

\newcommand{\putindex}[3]{\vtop{\hbox{\hspace{#3} $#1$}
            \hbox{\raise 6mm \hbox{$\scriptscriptstyle #2$}}}}

\newcommand{\gradx}[0]{\vtop{\hbox{\rm grad}
            \hbox{\raise 2.5mm \hbox{\rm \hspace{2mm} \footnotesize x}}}}

\newcommand{\grady}[0]{\vtop{\hbox{\rm grad}
            \hbox{\raise 2.5mm \hbox{\rm \hspace{2mm} \footnotesize y}}}}

\newcommand{\grad}[1]{\vtop{\hbox{\rm grad}
            \hbox{\raise 2.5mm \hbox{#1}}}}

\newcommand{\btb}{     \begin{tabbing}             }
\newcommand{\bte}{     \end{tabbing}               }

\newcommand{\ourmethodname}{OpenViGA}

\def\wacvPaperID{0000} 
\def\confName{arxiv}
\def\confYear{2025}

\title{\ourmethodname: Video Generation for Automotive Driving Scenes\\ by Streamlining and Fine-Tuning Open Source Models with Public Data }

\author{Björn Möller \quad Zhengyang Li  \quad Malte Stelzer \quad Thomas Graave\\ Fabian Bettels \quad Muaaz Ataya \quad Tim Fingscheidt \\
Technische Universität Braunschweig,
Institute for Communications Technology\\
{\tt\small \{bjoern.moeller, zhengyang.li, malte.stelzer, thomas.graave,}\\{\tt\small f.bettels, m.ataya, t.fingscheidt\}@tu-bs.de}
}
\begin{document}
\maketitle
\begin{abstract}
Recent successful video generation systems that predict and create realistic automotive driving scenes from short video inputs assign tokenization, future state prediction (world model), and video decoding to dedicated models.
These approaches often utilize large models that require significant training resources, offer limited insight into design choices, and lack publicly available code and datasets.
In this work, we address these deficiencies and present \ourmethodname, an open video generation system for automotive driving scenes. 
Our contributions are:
Unlike several earlier works for video generation, such as GAIA-1, we provide a deep analysis of the three components of our system by separate quantitative and qualitative evaluation: Image tokenizer, world model, video decoder.
Second, we purely build upon powerful pre-trained open source models from various domains, which we fine-tune by publicly available automotive data (BDD100K) on GPU hardware at academic scale.
Third, we build a coherent video generation system by streamlining interfaces of our components.
Fourth, due to public availability of the underlying models and data, we allow full reproducibility.
Finally, we also publish our code and models on Github\footnote{https://github.com/ifnspaml/OpenViGA}.
For an image size of 256x256 at 4 fps we are able to predict realistic driving scene videos frame-by-frame with only one frame of algorithmic latency.
\end{abstract}
\vspace{-0.3pt}
\section{Introduction} \label{sec:intro}
\begin{figure}
    \centering
    \includegraphics[width=0.49\textwidth]{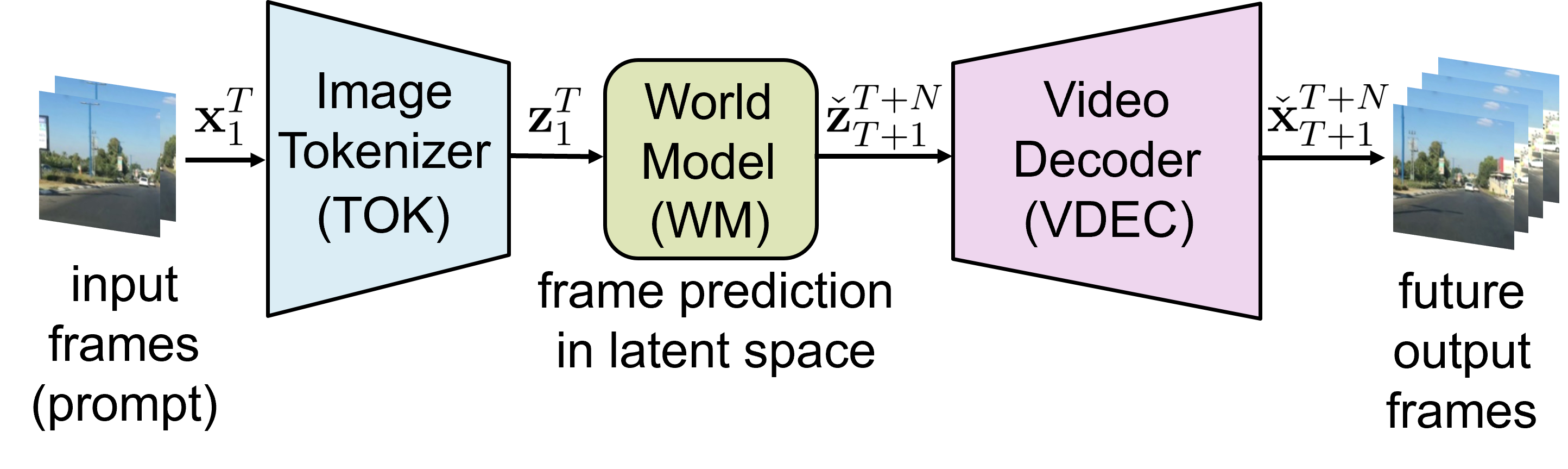}
    \caption{\textbf{Proposed video generation system}: It consists of an image tokenizer, encoding $T$ input frames $\mathbf{x}_1^T$ into a latent representation of discrete tokens $\mathbf{z}_1^T$, a world model then predicting $N$ subsequent latent image tokens $\check{\mathbf{z}}_{T+1}^{T+N}$, and a video decoder, generating $N$ output frames $\check{\mathbf{x}}_{T+1}^{T+N}$. Models are pre-trained, open-source, and then fine-tuned on public automotive data.}
    \label{fig:system_overview}
\end{figure}

Video generation models \cite{ho_video_2022} have improved significantly due to recent advances in large language models (LLMs) \cite{touvron_llama_2023-1, touvron_llama_2023} and generative AI \cite{ho_denoising_2020}, enabling the creation of realistic visual content from limited input data.
There is a growing focus on world model-based systems \cite{hu_gaia-1_2023,wu_ivideogpt_2024}, which learn structured representations of dynamic environments and enable the generation of long, temporally consistent video sequences through autoregressive next-token prediction.
Such systems typically operate in three stages: A tokenizer (TOK) encodes the input into discrete representations, a world model (WM) predicts a future token sequence, and finally a video decoder (VDEC) reconstructs the predicted token into video frames \cite{hu_gaia-1_2023}.

In autonomous driving, video generation offers a scalable approach to improve safety \cite{fingscheidt_deep_2022, houben_inspect_2022} by simulating diverse driving scenarios, including rare and harmful events such as evasive maneuvers, addressing the scarcity of critical training data \cite{zhao_drivedreamer-2_2024}.
Despite the demonstrated potential of video generation models such as GAIA-1 and \mbox{GAIA-2} \cite{hu_gaia-1_2023,russell_gaia-2_2025}, the lack of open-source implementations or training code, sparse description of architectures, and non-public datasets limit reproducibility.
In contrast, \textit{we propose \mbox{\ourmethodname{}}, an open video generation system for automotive driving scenes, see Fig.\ref{fig:system_overview}, based purely on open-source models and public data, allowing for full reproducibility.} 

Open-source foundation LLMs \cite{touvron_llama_2023-1, touvron_llama_2023}, having learned temporal prediction from vast corpora of text data, provide a powerful basis for WMs in video generation.
Extending them to multi-modal token prediction using image, video, and text data, enables them to model spatio-temporal structures.
However, training such a WM is demanding with respect to training time, dataset size, and computational resources.
Therefore, we build upon the pre-trained 7B parameter \texttt{LWM} \cite{liu_world_2024} model and employ low-rank adaptation (LoRA) \cite{hu_lora_2022} to efficiently fine-tune it.
For our image tokenizer and decoder, we also leverage a pre-trained \texttt{VQGAN} \cite{patil_amused_2024} model.
However, both these open-source models were originally trained on general-domain data causing deficiencies for automotive video generation.
Accordingly, we transfer them to the automotive domain by fine-tuning on public automotive data.

We utilize the large public driving dataset BDD100K \cite{yu_bdd100k_2020} to fine-tune our TOK on its images, and our VDEC and WM on its videos.
We also evaluate on the official image data splits.
To process the dataset's video content with limited GPU memory, we have to streamline all components towards a coherent system in terms of predicted token sequence length (WM), frame rate (WM), token per image (TOK), and input frame resolution (all).

In this paper, we first build a video generation system for automotive driving scenes (OpenViGA) and provide insights into its components by qualitative and quantitative evaluation.
We investigate loss architectures for image tokenizer and image decoder fine-tuning, evaluate the entire system, and investigate the world model's top-$k$ selection hyperparameter, controlling its creativity.
Second, we build upon general-purpose open-source models and fine-tune them on public automotive driving data, using limited GPU resources.
Third, we describe streamlining of individual open-source system components.
For full reproducibility, we publish our training and inference code.

\section{Related Work}\label{sec:rel_work}

\paragraph{Image Autoencoder Models}
Autoencoder (AE) models learn an efficient representation of data by encoding the input into a lower-dimensional representation and decoding the original data as accurately as possible. AEs are applied in areas such as image compression \cite{cheng_deep_2018}, anomaly detection \cite{zhou_anomaly_2017} and denoising \cite{vincent_extracting_2008}.
Incorporating convolutional layers \cite{masci_stacked_2011}, regularization techniques \cite{makhzani_adversarial_2016}, or probabilistic inference \cite{kingma_auto-encoding_2014}, image AEs are widely used to compress images into a compact latent representation for efficient processing in subsequent models.
Introducing quantization into the latent space \cite{van_den_oord_neural_2017, esser_taming_2021} results in discrete representations that enhance image generation and interpretability.
Following leading video generation approaches \cite{hu_gaia-1_2023, liu_world_2024}, we fine-tune a \texttt{VQGAN} \cite{esser_taming_2021} to encode patches of input frames into discrete tokens, which are then processed by our world model.

\paragraph{World Models}
World models (WMs) estimate missing information and predict future states of dynamic environments \cite{guan_world_2024}.
By learning latent representations of the environment, WMs can facilitate reinforcement learning \cite{zhang_storm_2023}, long-sequence predictions \cite{liu_world_2024}, and controllable video generation \cite{wu_ivideogpt_2024}.
In autonomous driving, WMs enhance navigation and enable environment interaction \cite{guan_world_2024}. 
\texttt{DriveGAN} \cite{kim_drivegan_2021} introduces an LSTM-based prediction model.
\texttt{DriverDreamer} \cite{wang_drivedreamer_2024}, \texttt{Drive-WM} \cite{wang_driving_2024}, and \texttt{GAIA-2} \cite{russell_gaia-2_2025} deploy diffusion models as WMs to predict multi-view videos, conditioned on frames, actions and additional inputs such as bounding boxes or maps.
\texttt{DriveDreamer-2} \cite{zhao_drivedreamer-2_2024} extends on this with an LLM \cite{touvron_llama_2023} for user-friendly trajectory planing.
\texttt{ADriver-I} \cite{jia_adriver-i_2023} also leverages an LLM to predict action priors for a video diffusion \cite{ho_denoising_2020} model. 
\texttt{GAIA-1} \cite{hu_gaia-1_2023} frames future prediction as next-token prediction over image patch tokens using an autoregressive transformer as WM, with a diffusion model for decoding, showing strong qualitative results.
However, the authors use a proprietary dataset and do not publish code or model weights, nor quantitative results beyond loss curves. 
\textit{We employ a pre-trained, open-source autoregressive transformer \cite{liu_world_2024} as our WM, adapt it to the automotive domain using public data, and, in contrast to prior works, publish our inference and training code as well as model weights.}

\paragraph{Video Generation Models}
Video generation models synthesize temporally consistent video sequences from input prompts. 
While GAN-based \cite{vondrick_generating_2016} and autoregressive models \cite{yan_videogpt_2021} show good results, recent video diffusion models (VDMs) \cite{ho_video_2022} further enhance temporal consistency and quality in generated videos. 
However, training such VDMs requires extensive hardware resources. \textit{Accordingly, we decide for a lower complexity attractive frame-by-frame approach by extending our fine-tuned \texttt{VQGAN} image decoder with temporal input context and fine-tune it on video data.}
\begin{figure*}[!ht]
    \centering
            \centering
            \includegraphics[width=1\textwidth]{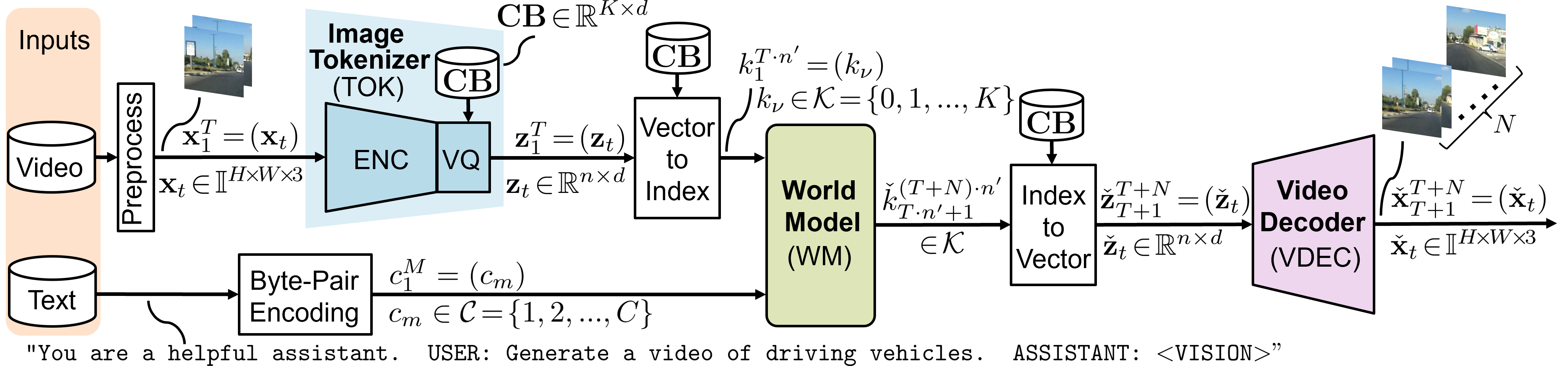}
            \caption{{\bf Inference} of the proposed {\bf video generation system}: A sequence of $N$ future frames $\check{\mathbf{x}}_{T+1}^{T+N}$ is predicted based on $T$ initial input images $\mathbf{x}_1^T$ and a fixed text prompt $c_1^M$. The tokenizer encodes the prompted image sequence $\mathbf{x}_1^T$ into a discrete token sequence $\mathbf{z}_1^T$. The world model predicts future tokens $\check{\mathbf{z}}_{T+1}^{T+N}$ for the video decoder to generate the corresponding image sequence $\check{\mathbf{x}}_{T+1}^{T+N}$. Details of the image tokenizer (TOK, i.e., ENC and VQ) are shown in Fig.\ \ref{fig:vqgan_train_architecture}. World model (WM) details are shown in Fig.\ \ref{fig:wm_train_architecture}.
            }
            \label{fig:system_inference}
\end{figure*}
\section{\ourmethodname}\label{sec:method}
Following Fig.\ \ref{fig:system_overview}, our proposed video generation system predicts $N$ future frames $\check{\mathbf{x}}_{T+1}^{T+N} \!=\! (\check{\mathbf{x}}_t)$, with $\mathbf{\check{x}}_t \in \mathbb{I}^{H\times W \times 3}$, of automotive driving scenes based on an input sequence $\mathbf{x}_1^{T} \! = \!(\mathbf{x}_t)$ of $T$ initial images, where $\VEC{x}_t \! \in  \! \mathbb{I}^{H \!\times\! W \!\times\! 3}$, with height $H$, width $W$, 3 color channels, and gray values normalized to the range $\mathbb{I}\!=\![-1,1]$. Also, a fixed text prompt is utilized, see Fig.\ \ref{fig:system_inference}. In the following, we describe the system components, the open-source models used, and our approach of streamlining and fine-tuning those models to automotive data.

\subsection{System Components}
\paragraph{Image Tokenizer}
The image tokenizer (TOK) in Fig.\ \ref{fig:system_inference} converts a preprocessed image $\mathbf{x}_{t}$ to a latent representation $\mathbf{z}_{t}\!=\!(\mathbf{z}_{t,\nu}) \in \mathbb{R}^{n \times d}$ of $n\!=\!16\times 16 \!=\! 256$ discrete tokens $\mathbf{z}_{t,\nu}$. It consists of an encoder (ENC), which encodes an image into $n$ latent space representations of an image patch, and a vector quantizer (VQ), which outputs token vectors $\mathbf{z}_{t,\nu} \in \mathbb{R}^d$ from a learnable codebook $\mathbf{CB} \in \mathbb{R}^{K \times d}$ with $K$ being the codebook size and codevector dimension $d$.
We deploy an ENC and VQ from a \texttt{VQGAN} \cite{esser_taming_2021} as image tokenizer (TOK).

\paragraph{World Model}
Based on prompted image tokens $\mathbf{z}_1^T \! = \! (\mathbf{z}_t)$ of $T$ initial frames and a text prompt $c_1^M \! = \! (c_m)$ consisting of $M$ text tokens, the world model (WM) predicts the discrete tokens of $N$ future frames, $\check{\mathbf{z}}_{T+1}^{T+N}\!=\!(\check{\mathbf{z}}_t)$.
Since the prompted image tokens $\mathbf{z}_t\!=\!(\mathbf{z}_{t,\nu})$ are discrete, a vector-index mapping converts $\mathbf{z}_t^T$ to a codebook index sequence $k_{1}^{T \cdot n^{\prime}}\!=\!(k_{\nu})$, with indices $k_{\nu} \in \mathcal{K} \!=\!\{0,...,K\}$, and patch position $\nu \in \{1,...,T \cdot n^{\prime}\}$ of these codebook vectors.
Also, end-of-image tokens ($k_{\tau \cdot n^{\prime}}\!=\!0, \quad\tau \in \{1,2,...,T\}$), with $n^{\prime}\!=\!n\!+\!1\!=\!257$, are inserted as structural information after each block of $n$ codebook indices that represent a prompted image.
Focusing on video continuation, we fix the text prompt to \texttt{"You are a helpful assistant. USER: Generate a video of driving vehicles. ASSISTANT: $<$VISION$>$}", which is byte-pair encoded to $M$ text tokens $c_1^M$.
Here, \texttt{$<$VISION$>$} marks the beginning of the codebook index sequence $k_{1}^{T \cdot n^{\prime}}\!=\!(k_{\nu})$ of $T$ tokenized initial ground-truth frames.
From its concatenated inputs ($c_1^M$, $k_{1}^{T \cdot n^{\prime}}$), the WM predicts $N$ future frames in an autoregressive loop, selecting the next image token in each of the $N \cdot n'$ total iterations.
The next image token is selected by first predicting a probability distribution and then sampling from the top-$k$ most probable entries.

\paragraph{Video Decoder}
The video decoder (VDEC) generates $N$ frames $\check{\mathbf{x}}_{T+1}^{T+N} \!=\! (\check{\mathbf{x}}_t)$ from the world model's temporal predictions $\check{\mathbf{z}}_{T+1}^{T+N} \!=\! (\check{\mathbf{z}}_t)$ one frame at a time, using a bidirectional input context window over three frames.
We deploy a 3D CNN as VDEC, derived from our fine-tuned 2D \texttt{VQGAN} image decoder (DEC) by 3D central inflation \cite{yu_magvit_2023}.

\subsection{Open Source Models}

\paragraph{VQGAN} 
The vector-quantized generative adversarial network (\texttt{VQGAN}) \cite{esser_taming_2021} integrates vector quantization \cite{van_den_oord_neural_2017}, autoencoders \cite{kingma_auto-encoding_2014}, and GANs \cite{goodfellow_generative_2014} to generate high-quality images with compact latent representations.
Our approach builds on an open-source general-purpose \texttt{VQGAN} model \cite{patil_amused_2024} originally trained for 2.5M steps on diverse real-world images at 256×256 resolution.
The TOK module (ENC + VQ) has 59.2M parameters, about 20\% of GAIA-1’s tokenizer, and the DEC block contains 87M.
\textit{However, deploying it for front-facing camera videos of automotive driving scenes, the model exhibits limited reconstruction quality.}

\begin{figure*}[!t]
    \centering
            \centering
            \includegraphics[width=0.99\textwidth]{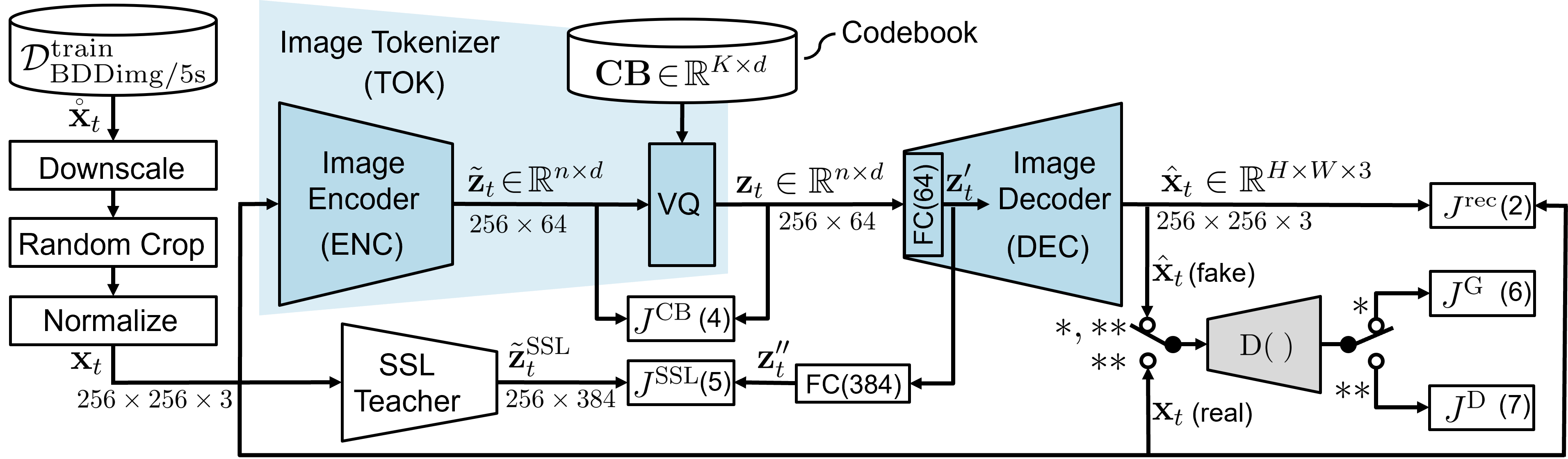}
            \caption{{\bf Fine-tuning} of the \texttt{VQGAN} {\bf image tokenizer} and {\bf image decoder} for a single training image $\mathbf{x}_t$ (selectors in positions $\ast$, as drawn). The image tokenizer consists of an encoder (ENC) and a vector quantizer (VQ) and is jointly trained with the image decoder (DEC), as part of a \texttt{VQGAN} model. ENC and DEC are optimized by a combination of reconstruction loss $J^{\mathrm{rec}}$, GAN loss (generator component $J^{\mathrm{G}}$) and self-supervised learning loss $J^{\mathrm{SSL}}$. The VQ's codebook $\mathbf{CB}$ and ENC are additionally optimized with a codebook loss $J^{\mathrm{CB}}$. The \textbf{discriminator} $\mathrm{D}(\:)$ is alternatingly optimized on the same batch (selector positions $\ast\ast$), using discriminator loss $J^{\mathrm{D}}$.}
            \label{fig:vqgan_train_architecture}
\end{figure*}
\paragraph{LWM}

The large world model (\texttt{LWM}) \cite{liu_world_2024} is a 7B-parameter multimodal autoregressive decoder-only transformer based on \texttt{LLaMA-2} \cite{touvron_llama_2023} capable of processing sequences up to 1M tokens of both video and text modalities with 32 decoder blocks.
As our WM, we adopt the open-source LWM-Chat-1M model, originally trained on text and four vision-language tasks with VQGAN-encoded inputs \cite{patil_amused_2024} and 495B tokens from diverse web datasets \cite{schuhmann_laion-5b_2022,bain_frozen_2021}. For details, see Supplement A.
{\it However, no dedicated driving video dataset containing recordings from forward-facing cameras was utilized.
Also, the task of predicting future latent frames based on initial latent frames was not addressed in \texttt{LWM} training.
Moreover, \texttt{LWM} was trained with 256 tokens per image defining the image tokenizer output size and thus limits the video frame resolution, which we have to streamline to fit to the BDD100K automotive dataset.}

\subsection{Our Streamlining and Fine-Tuning}
\label{sec:streamlining}
\paragraph{Streamlining}
Before separately fine-tuning the pre-trained models, we streamline the components within the system for processing automotive data with limited hardware resources.
While the \texttt{LWM} is capable of processing long token sequences yet with highly parallelized hardware, in practice, sequence length is limited by available GPU memory. 
The open-source \texttt{LWM-Chat-1M} was originally trained on video frames quantized to 256 tokens each and a frame rate of 4 fps, both of which we retain for our fine-tuning.
On a single \texttt{Nvidia A100} GPU with 80GB of VRAM, we were able to predict image tokens 4 seconds into the future during fine-tuning, yielding a total predicted sequence length of $N \cdot n^{\prime}\!=$ 4,112 tokens, which is the product of $N\! = 4$ fps$\cdot$ 4 $s\!=\!16$ frames, and $n^{\prime}\!=\!n+1\!=257$ image tokens including a manually inserted end-of-image token.
Given that automotive datasets typically provide frames at a resolution of $1.280 \times 720$ pixels, the 256-token per-frame constraint imposed by the world model presents a substantial challenge for tokenizing spatial features across the entire scene.
To address this, we first select an image tokenizer (as part of \texttt{VQGAN}) that allows for a relatively high input image size of $H \times W \times 3 \!=\! 256 \times 256 \times 3$ producing $\frac{H\!=\!256}{16}\times\frac{W\!=\!256}{16} \!=\!256$ image patch tokens.
To grasp a meaningful section of the automotive scene, we downscale the original BDD100K image $\mathring{\mathbf{x}}_t$ by a factor of 0.5 in height and width using bilinear interpolation \cite{Gonzales2008}.
This consequently increases the density of spatial detail, which adds difficulty in learning precise reconstructions during \texttt{VQGAN} fine-tuning.
The streamlined system takes a normalized image crop $\mathbf{x}_t$ of size $H\times W\times 3 \!=\! 256\times 256 \times 3$ from the image center as input, modeling the relevant portion of the driving scene.

Also, for later \texttt{VQGAN} fine-tuning (shown in Fig.\ \ref{fig:vqgan_train_architecture}), a teacher network is used to distill relevant information into the discrete tokens $\mathbf{z}_t$ by self-supervised learning (SSL).
To match the dimensions for the pre-trained \texttt{VQGAN}'s quantized representations $\mathbf{z}_t$ and the teacher's encoded latent representation $\tilde{\mathbf{z}}_t^{ \mathrm{SSL}}$, a fully connected adapter layer $\mathrm{FC(384)}$ is inserted.

\paragraph{Fine-Tuning the Tokenizer and Image Decoder}
We fine-tune the image tokenizer (TOK) jointly with the image decoder (DEC) as parts of a \texttt{VQGAN} model utilizing a combination of various loss functions, see Fig.~\ref{fig:vqgan_train_architecture}.
As the TOK compresses each image independently, we fine-tune on an \textit{image} training dataset $\mathcal{D}^{\mathrm{train}}_{\mathrm{BDDimg/5s}}$ (see Table \ref{tab:dataset_bdd100k}) and present the process for a single raw image $\mathring{\mathbf{x}}_t$, which is first downscaled, cropped, and normalized, yielding our ground-truth training image $\mathbf{x}_t$.

The TOK and DEC optimization can be expressed as total loss with hyperparameters $\lambda$:
\begin{equation}\label{eq:loss_total}
J^{\mathrm{total}}\!=\!J^{\mathrm{rec}} + \lambda^{\mathrm{CB}} J^{\mathrm{CB}} + \lambda^{\mathrm{SSL}}  J^{\mathrm{SSL}}  + \lambda^{\mathrm{G}} J^{\mathrm{G}}
.
\end{equation}

Our reconstruction loss combines the pixel-wise $L_1$, $L_2$, and the perceptual loss $J^{\prime}(\:)$ \cite{johnson_perceptual_2016} in a weighted sum
\begin{equation}\label{eq:loss_Jrec}
J^{\mathrm{rec}}(\hat{\mathbf{x}}_t, \mathbf{x}_t) =  
\lambda_{1}||\hat{\mathbf{x}}_t-\mathbf{x}_t||_1 +  \lambda_{2}||\hat{\mathbf{x}}_t -\mathbf{x}_t||_2^2 + 
\lambda^{\prime} J^{\prime}(\hat{\mathbf{x}}_t,\mathbf{x}_t).
\end{equation}
The perceptual loss encourages reconstructions $\hat{\mathbf{x}}_t$ that are perceptually similar to the reference image $\mathbf{x}_t$ by similar feature activation $\boldsymbol{\phi}_{\ell}(\hat{\mathbf{x}}_t)$ and $\boldsymbol{\phi}_{\ell}(\mathbf{x}_t)$ for an ImageNet-pre-trained loss network $\boldsymbol{\phi}$ with 
\begin{equation}\label{eq:loss_Jperc}
J^{\prime}(\hat{\mathbf{x}}_t , \mathbf{x}_t) = \sum_{\ell} \| \boldsymbol{\phi}_{\ell}(\hat{\mathbf{x}}_t) - \boldsymbol{\phi}_{\ell}(\mathbf{x}_t) \|_2^2,
\end{equation}
where $\ell$ refers to specific model layers \cite{johnson_perceptual_2016}.

The codebook $\mathrm{CB}$ is optimized using an embedding loss that utilizes the L\textsubscript{2} error to align the codebook vectors and the encoder output $\tilde{\mathbf{z}}$, while the commitment loss \cite{van_den_oord_neural_2017} forces the encoder to stick to a particular embedding, jointly forming the codebook loss
\begin{equation}\label{eq:loss_JCB}
J^{\mathrm{CB}}(\mathbf{z}_t , \tilde{\mathbf{z}}_t) = 
\underbrace{
\|\mathrm{sg}(\tilde{\mathbf{z}}_t) - \mathbf{z}_t\|_2^2 
}_{\text{embedding loss}}
+
\beta \cdot 
\underbrace{
\|\tilde{\mathbf{z}}_t - \mathrm{sg}(\mathbf{z}_t)\|_2^2 
}_{\text{commitment loss}}
,
\end{equation}
where $\mathrm{sg}(\cdot)$ denotes the stop-gradient operator and $\beta\!=\!0.25$ is a commitment weight.
As the quantization function (VQ) is non-differentiable, gradients of $J^{\mathrm{rec}}$,  $J^{\mathrm{SSL}}$, and $ J^{\mathrm{G}}$ are backwards copied from $\mathbf{z}_t$ to $\tilde{\mathbf{z}}_t$, influencing the codebook only indirectly \cite{esser_taming_2021}.

Also, a teacher model is leveraged to distill semantic meaning into the image patch tokens $\mathbf{z}_t$ by enforcing cosine similarity between a teacher's encoding $\tilde{\mathbf{z}}_t^{\mathrm{SSL}}$ and the quantized and transformed representation $\mathbf{z}^{\prime}_t$, serving as a self-supervised learning loss
\begin{equation}\label{eq:loss_JIB}
J^{\mathrm{SSL}}(\mathbf{z}^{\prime}_t , \mathbf{z}_t^\mathrm{SSL} )= 1 -  \frac{(\tilde{\mathbf{z}}^{\mathrm{SSL}}_t)^{\mathrm{T}} \cdot \mathbf{z}^{\prime\prime}_t}
{ \mathrm{max}( \|\tilde{\mathbf{z}}_t^{\mathrm{SSL}}\|_2 \cdot\|\mathbf{z}^{\prime\prime}_t\|_2, \epsilon)},
\end{equation}
where $\epsilon \!=\! 10^{-8}$ and $\mathbf{z}^{\prime\prime}_t \in \mathbb{R}^{384}$ is the output of a learnable fully connected layer with input $\mathbf{z}^{\prime}_t \in \mathbb{R}^{64}$. 
We adopt \texttt{ViT-s DINOv2} \cite{oquab_dinov2_2024} as the teacher, extending GAIA-1's use of \texttt{DINOv1} \cite{caron_emerging_2021}, with experimental validation.

To generate realistic-looking images, a GAN loss approach is adopted, consisting of two loss components with complementary objectives for discriminator $\mathrm{D(\:)}$ and generator (TOK+DEC).
Included in (\ref{eq:loss_total}) is the generator loss
\begin{equation}\label{eq:loss_GAN}
J^{\mathrm{G}}(\hat{\mathbf{x}}_t) = -\mathrm{D}(\hat{\mathbf{x}}_t),
\end{equation}
depicted with selector positions $\ast$ in Fig.\ \ref{fig:vqgan_train_architecture}, which optimizes the generator (TOK+DEC) to produce fake images, misleading the discriminator $\mathrm{D(\:)}$ into classifying them as real.

Complementing the total loss $J^{\mathrm{total}}$ (\ref{eq:loss_total}), a patch-based discriminator $\mathrm{D}(\:)$ is trained from scratch to distinguish real and generated images using a separate discriminator loss
\begin{equation}\label{eq:loss_Jdisc}
J^{\mathrm{D}}(\hat{\mathbf{x}}_t , \mathbf{x}_t) = 0.5(\mathrm{ReLU}(1-\mathrm{D}(\mathbf{x}_t)) + 
\mathrm{ReLU}(1+\mathrm{D}(\hat{\mathbf{x}}_t)))
,
\end{equation}
encouraging $\mathrm{D}(\mathbf{x}_t) \approx 1$ for real samples and $\mathrm{D}(\hat{\mathbf{x}}_t) \approx -1$ for fake (i.e., auto-encoded) samples, see Fig.\ \ref{fig:vqgan_train_architecture} (selector positions $\ast \ast$).
We adapt the patch-based discriminator's architecture from Esser et al.\ \cite{esser_taming_2021}, reducing the number of patches to 64, validated empirically ("our $\mathrm{D}(\;)$").

\begin{figure}[!t]
    \centering
            \centering
            \includegraphics[width=.49\textwidth]{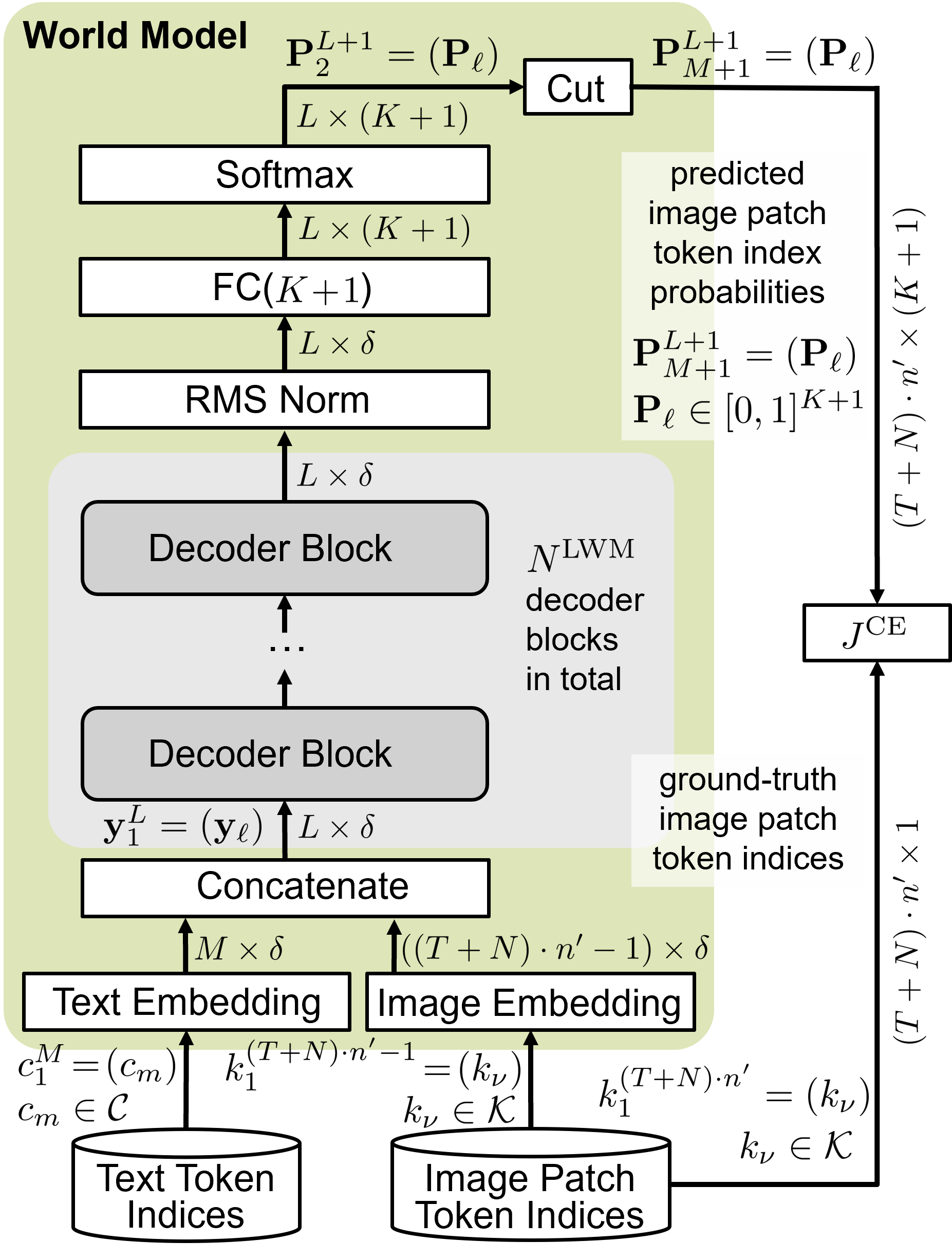}
            \caption{{\bf Fine-tuning} of the {\bf world model} for a single training sample consisting of a text index sequence $c_1^M$ and an image token index sequence $k_1^{(T+ N) \cdot n'}$. The cross-entropy loss $J^{\mathrm{CE}}$ is calculated between predicted probability distributions $\mathbf{P}_{M+1}^{L+1}\!=\!(\mathbf{P}_{\ell})$ for image token indices and ground-truth indices $k_1^{(T + N) \cdot n^{\prime}}$.}
            \label{fig:wm_train_architecture}
\end{figure}
\paragraph{Fine-Tuning the World Model}
We fine-tune the world model (WM) on a 4 fps \textit{video} training dataset $\mathcal{D}_{\mathrm{BDDvid-4fps}}^{\mathrm{train-70k}}$ (cf.\ Table \ref{tab:dataset_bdd100k}), with each sample consisting of text token index sequence $c_1^M\!=\!(c_m)$ and an image patch token index sequence $k_1^{ (T+ N) \cdot n'} \!=\!(k_{\nu})$, where each of the $T\!=\!2$ initial and $N\!=\!14$ future video frames is represented by $n'\!=\!257$ indices, see Fig.\ \ref{fig:wm_train_architecture}.
The input sequence $\mathbf{y}_1^L\!=\!(\mathbf{y}_\ell)$, formed by embedding and concatenating $c_1^M$ and $k_1^{(T+ N) \cdot n'-1} \!=\!(k_{\nu})$, is processed by $N^{\mathrm{LWM}}\!=\!32$ decoder blocks under causal attention to predict the next image token index. Fine-tuning uses teacher forcing and cross-entropy loss over the predicted probability distributions $\mathbf{P}_{M+1}^{L+1}\!=\!(\mathbf{P}_\ell)$ for all target image patch tokens $k_{\nu}$.
Due to the size of the WM, full fine-tuning is impractical, so we use the parameter-efficient LoRA \cite{hu_lora_2022} method, fine-tuning only a small set of additional adaptor weights for all linear and embedding layers, with full fine-tuning limited to normalization layer weights.
This results in 6.97B total model parameters, with only 2.39\% of these being fine-tuneable.
To further reduce memory usage, frozen parameters are stored in the lower-precision format \texttt{bfloat16}, while only fine-tunable parameters remain in \texttt{FP32}. More details in Supplement B.

\paragraph{Fine-Tuning the Video Decoder}
We fine-tune our video decoder (VDEC) on the 4 fps \textit{video} training dataset $\mathcal{D}_{\mathrm{BDDvid-4fps}}^{\mathrm{train-70k}}$ (cf.\ Table \ref{tab:dataset_bdd100k}), using samples of 3 consecutive frames $\mathbf{x}_{t-1}^{t+1}$. 
The frames are preprocessed and then tokenized separately using the fine-tuned and then frozen TOK, and jointly decoded to $\hat{\mathbf{x}}_{t-1}^{t+1}$.
For optimization, we employ the image reconstruction loss $J^{\mathrm{rec}}$ (\ref{eq:loss_Jrec}) and the generator component $J^{\mathrm{G-3D}}$ of a sequence-based GAN loss in analogy to (\ref{eq:loss_GAN}), as video decoder loss
\begin{equation}\label{eq:loss_JVDEC}
\begin{split}
J^{\mathrm{VDEC}}(\hat{\mathbf{x}}_{t-1}^{t+1},\mathbf{x}_{t-1}^{t+1})\!=\! \sum_{\tau\!=\!t-1}^{t+1}J^{\mathrm{rec}}(\hat{\mathbf{x}}_{\tau},\mathbf{x}_{\tau}) \qquad  \\  + \: J^{\mathrm{G-3D}}(\hat{\mathbf{x}}_{t-1}^{t+1})
.
\end{split}
\end{equation}

Consequently, we use the discriminator loss component $J^{\mathrm{D-3D}}(\hat{\mathbf{x}}_{t-1}^{t+1},\mathbf{x}_{t-1}^{t+1})$ after (\ref{eq:loss_Jdisc}) to train a 3D patch-based discriminator $\mathrm{D}^{\mathrm{VDEC}}(\;)$ from scratch, which jointly classifies sequences of three reconstructed frames $\hat{\mathbf{x}}_{t-1}^{t+1}$. We adopt our $\mathrm{D}(\;)$ architecture from TOK+DEC training and inflate it to 3D, inspired by Yu et al. \cite{yu_magvit_2023}.
\begin{table}[!tb]
	\centering
	\caption{\textbf{Amounts of videos and images} used in this work. 
 }
	\begin{tabular}{l @{\hspace{5pt}} c@{\hspace{8pt}}c@{\hspace{
    8pt}}c@{\hspace{8pt}}c@{\hspace{8pt}}c}
		\toprule
	 Notation  & Data format & Fps & \#train & \#val & \#test \\
     \midrule
        $\mathcal{D}_{\mathrm{BDDvid}}$ & video &30& 70k & 10k & 20k \\
        $\mathcal{D}_{\mathrm{BDDvid-4fps}}^{\mathrm{train-70k}}$ &video & 4&  70k & - & - \\
        $\mathcal{D}_{\mathrm{BDDvid-4fps}}^{\mathrm{val-500}}$ &video & 4& - & 500 & - \\
        \midrule
        $\mathcal{D}_{\mathrm{BDDimg}}$ & images &-& 70k & 10k & 20k \\
         $\mathcal{D}_{\mathrm{BDDimg/5s}}$ & images &-& $\sim$ 538k & - & - \\
	  \bottomrule
	\end{tabular}
	\label{tab:dataset_bdd100k}
\end{table}
\section{Experiments and Discussion}\label{sec:experiments}
\subsection{Experimental Setup}

\paragraph*{Dataset}\label{sec:data}
The Berkeley Deep Drive Dataset (BDD100K) \cite{yu_bdd100k_2020} is the largest open source driving video dataset containing 100,000 videos, most of them with a duration of 40 seconds.
These have been recorded in various conditions, such as weather, lighting, and driving routes, captured at 30 fps with image size $\mathring{H} \times \mathring{W}\!=\!720 \times 1,280$.
As detailed in Table \ref{tab:dataset_bdd100k}, the BDD100K video dataset $\mathcal{D}_{\mathrm{BDDvid}}$ comprises a training, validation, and test split, with 70k, 10k, and 20k videos, respectively.
For WM and VDEC fine-tuning and validation, we create a subset of 70k training videos $\mathcal{D}^{\mathrm{train-70k}}_{\mathrm{BDDvid-4fps}}$
and a subset of 500 validation videos, $\mathcal{D}^{\mathrm{val-500}}_{\mathrm{BDDvid-4fps}}$, both subsampled to 4 fps.
BDD100K images $\mathcal{D}_{\mathrm{BDDimg}}$, a subset of 
$\mathcal{D}_{\mathrm{BDDvid}}$, contains one frame per video extracted just after 10 seconds.
For fine-tuning the image TOK (and image DEC), we created a larger custom image subset $\mathcal{D}^{\mathrm{train}}_{\mathrm{BDDimg/5s}}$ with about 538k training images, sampled at 0.2 fps from $\mathcal{D}^{\mathrm{train}}_{\mathrm{BDDvid}}$, while we still validate on the official validation split $\mathcal{D}^{\mathrm{val}}_{\mathrm{BDDimg}}$.
The BDD100K video training set $\mathcal{D}^{\mathrm{train}}_{\mathrm{BDDvid}}$ provides 778 hours of training data \textit{with approximately 84M unique images, which is only about 15\% of the 5,100 hours of GAIA-1's unpublished dataset with about 420M unique images} \cite{hu_gaia-1_2023}.

\paragraph*{Metrics}
To assess image reconstruction  quality, we use PSNR \cite{Salomon2004}, SSIM \cite{wang_image_2004}, MS-SSIM \cite{wang_multiscale_2003}, and LPIPS \cite{zhang_unreasonable_2018}, comparing decoded images to ground-truth references. 
For \textit{generated} image/video frames, direct ground-truth comparison is not reasonable, accordingly, we report the Fréchet inception distance (FID) \cite{heusel_gans_2017}, the CLIP maximum mean discrepancy (CMMD) \cite{jayasumana_rethinking_2024}, and for video data also the Fréchet video distance (FVD) \cite{unterthiner_fvd_2019}.
The number of evaluated video frames is noted in the subscript (e.g., FVD\textsubscript{14}).
These perceptual metrics enable quality assessment of the WM and image/video decoder by evaluating predicted video frames while allowing them to diverge from corresponding ground-truth frames.
Thereby, we provide a \textit{quantitative} analysis of image/video quality, while \mbox{GAIA-1} \cite{hu_gaia-1_2023} authors and \mbox{GAIA-2} \cite{russell_gaia-2_2025} authors only report qualitative results.

\paragraph*{Training Details}
All our models are fine-tuned using the AdamW optimizer, distributing the batch across four \texttt{Nvidia H100} 94GB GPUs.
Our TOK+DEC (\texttt{VQGAN}) is fine-tuned for 200k steps with a batch size of 80, comprising 2k steps of linear warm-up to an initial learning rate of $5 \times 10^{-5}$ and 150k steps of cosine decay to a final learning rate of $5 \times 10^{-7}$, \textit{resulting in only 24 hours of fine-tuning}. 
The discriminator loss $J^{D}$ (\ref{eq:loss_Jdisc}) and generator loss $J^{\mathrm{G}}$ (\ref{eq:loss_GAN}) are applied after 20k steps.
Our WM is fine-tuned for 28.3k steps with a batch size of 24, comprising 250 steps of linear warm-up to an initial learning rate of $6 \times 10^{-4}$ and 15k steps of cosine decay to a final learning rate of $6 \times 10^{-5}$, \textit{resulting in 65 hours of fine-tuning}. Other configurations follow LLaMA-2 \cite{touvron_llama_2023}.
Our VDEC is fine-tuned for 100k steps with a batch size of 48, comprising 100 steps of linear warm-up to an initial learning rate of $5 \times 10^{-5}$ followed by a cosine decay schedule to $5 \times 10^{-7}$, \textit{resulting in 65 hours of fine-tuning.}
The discriminator loss $J^{D-3D}$ and generator loss $J^{\mathrm{G-3D}}$  are applied after 2k steps.
All code is based on the \texttt{PyTorch}\cite{paszke_pytorch_2019} and \texttt{jax}\cite{jax2018github} frameworks.

\subsection{Results and Discussion}
\begin{table*}[!tb]
	\centering
	\caption{\textbf{Tokenizer} and {\bf image decoder loss function} ablations: Quantitative results for \textit{omitting} (-$J$) loss components during fine-tuning with the total loss $J^{\mathrm{total}}$ (\ref{eq:loss_total}), employing $J^{D}$(\ref{eq:loss_Jdisc}) for discriminator training, with $\mathrm{D}(\:)$ from \cite{esser_taming_2021} and $J^{\mathrm{SSL}}$ (\ref{eq:loss_JIB}) using \texttt{DINOv1} \cite{caron_emerging_2021},  evaluated on $\mathcal{D}_{\mathrm{BDDimg}}^{\mathrm{val}}$. 
    \texttt{VQGAN} models (TOK + DEC) were fine-tuned on $\mathcal{D}_{\mathrm{BDDimg/5s}}^{\mathrm{train}}$ for 200k iterations. Weights of omitted losses are set to zero. Visual examples for all experiments are shown in Supplement C.
    Best results are in bold font, second best underlined. PSNR in (dB).
 }
	\begin{tabular}{l@{\hskip 2pt}|cccccc}
		\toprule
         Loss functions  & PSNR $\uparrow$	 & SSIM $\uparrow$ & MS-SSIM $\uparrow$ & LPIPS $\downarrow$ & FID  $\downarrow$ & CMMD $\downarrow$ \\
        \midrule
        $J^\mathrm{total}$ (\ref{eq:loss_total}),$J^{\mathrm{D}}$ (\ref{eq:loss_Jdisc}) & 25.75 & 0.7630 & 0.9022  & \underline{0.1170} & \textbf{5.48} & \textbf{0.074}  \\
        - $J^{\mathrm{SSL}}$ (\ref{eq:loss_total}), (\ref{eq:loss_JIB}) & \underline{26.31} & \underline{0.7794} & \underline{0.9140} & \textbf{0.1096} & \underline{5.79} & 0.122 \\ 
        - $J^{\prime}$ of $J^{\mathrm{rec}}$ (\ref{eq:loss_total}), (\ref{eq:loss_Jrec}), (\ref{eq:loss_Jperc}) & 25.75 & 0.7487 & 0.8943 & 0.1756 & 18.64 & 0.594  \\
        - $J^{\mathrm{L2}}$ of $J^{\mathrm{rec}}$ (\ref{eq:loss_total}), (\ref{eq:loss_Jrec}) & 23.87 & 0.7304 & 0.8713 & 0.1307 & 6.10  & \underline{0.107} \\
        - $J^{\mathrm{G}}$ (\ref{eq:loss_total}), (\ref{eq:loss_GAN}) & \textbf{27.11} & \textbf{0.8041} & \textbf{0.9178} & 0.1759 & 17.97 & 0.393 \\
        \midrule
        No fine-tuning & 25.08 & 0.7690 & 0.9018  & 0.1207 & 5.82  & 0.385 \\ 
	  \bottomrule
	\end{tabular}
	\label{tab:vqgan_loss_ablations}
\end{table*}
\paragraph*{Image Tokenizer}
Table \ref{tab:vqgan_loss_ablations} shows the results of fine-tuning the pre-trained \texttt{VQGAN} (TOK+DEC) using total loss $J^{\mathrm{total}}$ (\ref{eq:loss_total}) and discriminator loss $J^{\mathrm{D}}$ (\ref{eq:loss_Jdisc}), and explores the effects of individual loss components by omitting them ($-J$) during fine-tuning. 
Although we employ a smaller tokenizer and partly different loss definitions, we start our investigations with the exact GAIA-1 loss weights \cite{hu_gaia-1_2023}: $\lambda^{\mathrm{L1}}\!=\!0.2$, $\lambda^{\mathrm{L2}}\!=\! 2.0$,  $\lambda^{\prime}\!=\!0.1$, $\lambda^{\mathrm{G}}\!=\!1.0$, $\lambda^{\mathrm{CB}}\!=\!1.0$, $\lambda^{\mathrm{SSL}}\!=\!0.1$. 

\begin{table*}[!t]
	\centering
	\caption{\textbf{Tokenizer} and {\bf image decoder discriminator and loss weight} ablations: Quantitative results for using our $\mathrm{D}(\:)$ with varying weights $\lambda^{\prime}$, $\lambda^\mathrm{G}$ for perceptual loss and generator loss during fine-tuning with the total loss $J^{\mathrm{total}}$ (\ref{eq:loss_total}), employing $J^{D}$(\ref{eq:loss_Jdisc}) for discriminator training, with $J^{\mathrm{SSL}}$ (\ref{eq:loss_JIB}) using \texttt{DINOv1} \cite{caron_emerging_2021}, evaluated on $\mathcal{D}_{\mathrm{BDDimg}}^{\mathrm{val}}$.  \texttt{VQGAN} models (TOK + DEC) were fine-tuned on  $\mathcal{D}_{\mathrm{BDDimg/5s}}^{\mathrm{train}}$ for 200k iterations.  Best results are in bold font, second best underlined. PSNR in (dB).
 }
	\begin{tabular}{lcc|c|cccccc}
		\toprule
        Fine-tuning with $\mathrm{D}(\:)$ & $\lambda^{\prime}$ &    $\lambda^{G} \!$&  $\frac{\lambda^{\mathrm{G}}}{\lambda^{\prime}}$  & PSNR $\uparrow$	 & SSIM $\uparrow$ & MS-SSIM $\uparrow$ & LPIPS $\downarrow$ & FID  $\downarrow$ & CMMD $\downarrow$ \\
        \midrule
        $\mathrm{D}(\:)$ from \cite{esser_taming_2021} & 0.1 & 1.0  & 10 & 25.75 & 0.7630 & 0.9022  & 0.1170 & 5.48 & 0.074  \\
        \midrule
       Our $\mathrm{D}(\:)$ (Sec.\ \ref{sec:streamlining}) & 0.1 & 1.0  & 10 & \underline{25.88} & 0.7685  & 0.9039 & 0.1148 & 5.14 & 0.065  \\
     \hspace{4em}.  & 0.1 & 1.5   & 15 & \textbf{25.89} & 0.7694 & 0.9044 & 0.1146 & 5.32 & 0.064 \\
    \hspace{4em}. & 0.3 & 1.0  & 3.3 & 25.68       & \underline{0.7704}    & \underline{0.9049}   & 0.1050 & 4.43 & 0.053 \\
    \hspace{4em}.  & 0.3 & 1.2 & 4 & 25.68 & \textbf{0.7712} & \textbf{0.9050} & 0.1053 & 4.45 & 0.059 \\
    \hspace{4em} (\textbf{proposed}) & 1.0 & 1.0 & 1 & 25.14 & 0.7649 & 0.8991 & \textbf{0.1035} & 3.97 & \textbf{0.048} \\ 
    & 2.0 & 1.0 & 0.5 & 24.74 & 0.7616 & 0.8941 & 0.1042 & \textbf{3.55} & \underline{0.050} \\
    & 2.0 & 2.0 & 1 &  24.77 & 0.7623 & 0.8954 & \underline{0.1040} & \underline{3.93} & 0.051 \\
      \bottomrule
	\end{tabular}
	\label{tab:vqgan_discriminator_ablations}
\end{table*}
Omitting  $J^{\mathrm{SSL}}$ improves four metrics, but on the important perceptual FID and particularly CMMD it falls behind $J^{\mathrm{total}}$. Removing the perceptual contribution $J^{\prime}$ of $J^{\mathrm{total}}$ expectedly leads to catastrophic perceptual metrics, which are also not improved vs.\ $J^{\mathrm{total}}$ by omitting $J^{\mathrm{L2}}$. Due to the strong PSNR, SSIM, MS-SSIM, one might be tempted to omit $J^{\mathrm{G}}$, but the price is again significantly worse perceptual metrics.
For all subsequent investigations, we adopt the total loss $J^{\mathrm{total}}$ (\ref{eq:loss_total}), as it overall yields best perceptual performance (LPIPS, FID, CMMD).
\def\imgspace{0.1}
\begin{figure}[t!]
\centering
    \begin{subfigure}[t]{0.245\linewidth}
        \centering
        \includegraphics[width=\textwidth]{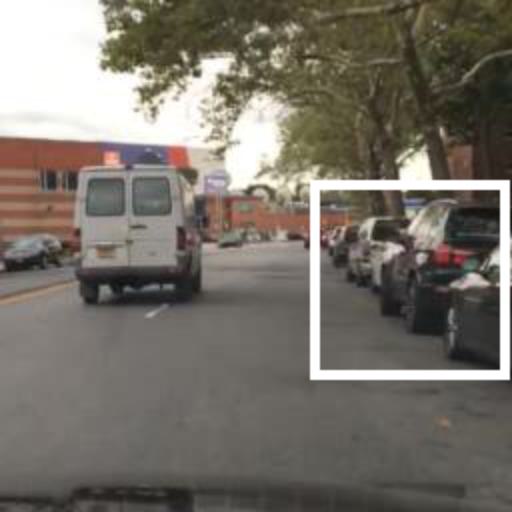}
        \caption*{a) Orig. image $\mathbf{x}_t$}
        \label{fig:tokenizer_vis_a}
    \end{subfigure}%
    \hfill
    \begin{subfigure}[t]{0.245\linewidth}
        \centering
        \includegraphics[width=\textwidth]{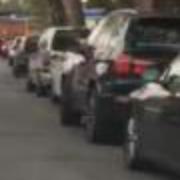}
        \caption*{b) Orig. RoI}
        \label{fig:tokenizer_vis_b}
    \end{subfigure}%
    \hfill
    \begin{subfigure}[t]{0.245\linewidth}
        \centering
        \includegraphics[width=\textwidth]{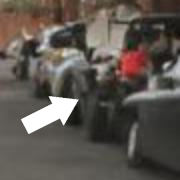}
        \caption*{c) No fine-tuning}
        \label{fig:tokenizer_vis_c}
    \end{subfigure}%
    \hfill
     \begin{subfigure}[t]{0.245\linewidth}
        \centering
        \includegraphics[width=\textwidth]{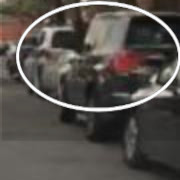}
        \caption*{d) \textbf{proposed}}
        \label{fig:tokenizer_vis_e}
    \end{subfigure}%
\caption{
\textbf{Example} of a region of interest (RoI) for a transcoded (ENC+VQ+DEC = TOK+DEC) image from $\mathcal{D}_{\mathrm{BDDimg}}^{\mathrm{val}}$.
(a) The RoI is marked in the original image $\mathbf{x}_t$, and (b) magnified for method comparison.
(c) Without domain-specific fine-tuning, the pre-trained \texttt{VQGAN} produces poor reconstructions of automotive objects (here: parked cars).
(d) Proposed fine-tuning significantly improves image quality.
Best viewed on screen.
}
\label{fig:tokenizer_quali}
\end{figure}

In Table \ref{tab:vqgan_discriminator_ablations}, 
we analyze and optimize the loss weights $\lambda'$ and $\lambda^{\mathrm{G}}$ of the important $J^{\mathrm{total}}$ contributions $J'$ and $J^{\mathrm{G}}$, respectively. We particularly show fine-tuning results using the discriminator from \cite{esser_taming_2021}, and in more detail {\it our} discriminator $\mathrm{D}(\:)$ (cf.\ Sec.\ \ref{sec:streamlining}) modified to classify fewer patches, each with an increased receptive field.
We finally propose and further use the TOK+DEC fine-tuning configuration of $J^{\mathrm{total}}$ (\ref{eq:loss_total}), our $\mathrm{D}(\:)$, and $\lambda^{\prime}\!=\!1$, $\lambda^{\mathrm{G}}\!=\!1$, as it is overall strongest in the perceptual metrics and therefore shows highest potential for application in a generative time-predictive framework as ours.

Fig.\ \ref{fig:tokenizer_quali} shows successful TOK+DEC fine-tuning, presenting an original image of an automotive scene (Fig.\ \ref{fig:tokenizer_quali}a), a zoomed-in region of interest highlighting parked cars (Fig.\ \ref{fig:tokenizer_quali}b), and corresponding transcoded model outputs (\ref{fig:tokenizer_quali}c,d). The non-fine-tuned \texttt{VQGAN} (TOK+DEC) exhibits poor reconstructions of car parts (Fig.\ \ref{fig:tokenizer_quali}c).
Fine-tuning with our proposed configuration from Table \ref{tab:vqgan_discriminator_ablations} restores structure and details, thereby improving perceptual quality.
More ablation results and visual examples are in Supplement C.

\paragraph*{World Model}
Table \ref{tab:teacher_ablations} presents the overall system performance with WM, but still using the image decoder DEC. Different teacher models are used for self-supervised learning loss $J^{\mathrm{SSL}}$ (\ref{eq:loss_JIB}) during TOK + DEC fine-tuning. Evaluation is performed on video data $\mathcal{D}_{\mathrm{BDDvid}}^{\mathrm{val-500}}$ for $N\!=\!14$ predicted frames based on $T\!=\!2$ conditioning frames.
While GAIA-1 uses a probably larger \texttt{DINO} (Version 1) for $J^{\mathrm{SSL}}$, our best FVD score is achieved with $J^{\mathrm{SSL-DINOv2}}$. This demonstrates the effectiveness of \texttt{DINOv2}'s representations over \texttt{DINO} or no $J^{\mathrm{SSL}}$ guidance, motivating its use as the teacher for our finally proposed TOK+DEC fine-tuning, therefore used in all consecutive experiments.
\begin{table}[!t]
	\centering
	\caption{\textbf{Distillation teacher model} ablations: Quantitative results for different teacher models used with $J^\mathrm{SSL}$ during fine-tuning of the tokenizer (TOK) and image decoder (DEC). Results were calculated for $N\!=\!14$ on the $\mathcal{D}_{\mathrm{BDDvid}}^{\mathrm{val-500}}$ subset. 
    Best results are in bold font.
    Top-$k$ value provided.
 }
	\begin{tabular}{l@{\hskip 2pt}|c|ccc}
		\toprule
        TOK/DEC  & \multicolumn{4}{c}{Validation: TOK/WM/DEC} \\
	fine-tuned with & $k$ & FID\textsubscript{14} $\downarrow$\!\! & CMMD\textsubscript{14} $\downarrow$\!\! & FVD\textsubscript{14} $\downarrow$ \\
        \midrule 
        No $J^{\mathrm{SSL}}$ & 50  & 14.65 & 0.105 & 188.59\\
        $J^{\mathrm{SSL-DINO}}$ & 50 & \textbf{14.25} & 0.099 & 181.55 \\
        $J^{\mathrm{SSL-DINOv2}}$ & 50 & 14.72 & \textbf{0.083} & \textbf{178.97}\\
	  \bottomrule
	\end{tabular}
	\label{tab:teacher_ablations}
\end{table}
\begin{table}[!t]
    \centering
    \caption{\textbf{World model top-k} ablations: Quantitative results for various configurations of top-$k$ sampling on the $\mathcal{D}_{\mathrm{BDDvid}}^{\mathrm{val-500}}$ subset with  $N\!=\!14$, $J^{\mathrm{SSL-DINOv2}}$ used for DEC optimization; VDEC builds upon that. Best results are in bold font.
    Top-$k$ value provided.
 }
	\begin{tabular}{l@{\hskip 2pt}|@{\hskip 2pt}c@{\hskip 2pt}|@{\hskip 2pt}c@{\hskip 4pt}c@{\hskip 4pt}c@{\hskip 0pt}}
		\toprule
		System & $k$ & FID\textsubscript{14} $\downarrow$ & CMMD\textsubscript{14} $\downarrow$ & FVD\textsubscript{14} $\downarrow$  \\
        \midrule
        TOK+DEC & - & 7.19 & 0.048 & 104.35 \\
        \midrule 
        \multirow{6}{*}{TOK+WM+DEC}
        & 1 & 28.04 & 0.255 & 644.71\\
        & 5 & 20.26 &  0.114 & 296.95 \\
        & 10 & 18.23 & 0.092 & 243.97\\
        & 50 & 14.72 & 0.083 & 178.97 \\
        & 200 & 12.22 & \textbf{0.081} & \textbf{160.48} \\
        & 1000 & \textbf{11.29} & 0.090 & 163.80 \\
	\midrule
        TOK+VDEC & - & 8.85 & 0.155 & 74.29\\
        \midrule 
        \multirow{6}{*}{\shortstack{OpenViGA \\ (TOK+WM+VDEC)}} 
        & 1 & 28.61 & 0.411 & 646.03 \\
        & 5 & 22.14 &0.244 & 276.93\\
        & 10 & 19.94 & \textbf{0.215} & 210.71\\
        & 50 & 16.67 & 0.229 & 153.19 \\
        & 200 & 14.22 &  0.241 & 136.41\\
        & 1000 & \textbf{13.29} & 0.248 & \textbf{132.16}\\
      \bottomrule
	\end{tabular}
	\label{tab:lwm_ablations}
\end{table}

In Table \ref{tab:lwm_ablations}, we investigate the world model's performance depending on the top-$k$ hyperparameter, which controls the breadth of token selection during inference.
We show TOK+DEC and TOK+VDEC as reference, which perform no prediction (i.e., no WM), instead, here, we only transcode the video frames.
VDEC is based on DEC trained with $J^{\mathrm{SSL-DINOv2}}$ and our $\mathrm{D}(\;)$.
We observe that temporal prediction by the WM of course harms the perceptual metrics, but values in the range $k=10 ... 1000$ improve the metrics. For our finally proposed OpenViGA system, we recommend $k=1000$ as this choice provides the best perceptive video quality (FVD\textsubscript{14}).
Visual examples of generated videos are given in the Supplement D.

\paragraph*{Image/Video Decoder}
Table \ref{tab:VDEC} presents our final quantitative comparison of the image decoder (DEC) and the video decoder (VDEC) for temporal predictive video generation using the WM.
We compare results of both systems using their strongest top-$k$ results in terms of FVD\textsubscript{14}.
We observe that using our VDEC yields superior FVD\textsubscript{14} performance compared to our DEC (132.16 vs. 160.48), justifying its adoption in our finally proposed OpenViGA system.
\section{Limitations}\label{sec:limitations}
Our 3D CNN video decoder, trained with GAN and reconstruction losses, likely produces lower-quality frames than modern diffusion-based methods. However, it offers a simple-to-train solution that effectively handles quantized latent image patch tokens and seamlessly integrates into our system. While also being effective in its frame-by-frame processing, extending its temporal context might allow to enhance visual coherence beyond our current achievements.

\begin{table}[t]
	\centering
	\caption{\textbf{Image/video decoder} and \textbf{final results with strongest top-$k$}: Quantitative results for deploying the \textit{image} decoder (DEC) or \textit{video} decoder (VDEC) for  generating video frames from WM output. Results were calculated for $N\!=\!14$ on the $\mathcal{D}_{\mathrm{BDDvid}}^{\mathrm{val-500}}$ subset. VDEC is based on DEC trained with $J^{\mathrm{SSL-DINOv2}}$. \ourmethodname{} employs TOK/WM/VDEC. Top-$k$ value provided.
    Best results are in bold font.
 }
	\begin{tabular}{l|@{\hskip 2pt}c@{\hskip 2pt}|@{\hskip 4pt}c@{\hskip 4pt}c@{\hskip 4pt}c}
	\toprule
	System \!\! & $k$ & FID\textsubscript{14} $\downarrow$ & CMMD\textsubscript{14} $\downarrow$ & FVD\textsubscript{14} $\downarrow$ \\
        \midrule
        TOK+WM+DEC & \phantom{0}200 & 12.22 & \textbf{0.081} & 160.48  \\
        \ourmethodname{} & 1000 & \textbf{13.29} & 0.248 & \textbf{132.16}\\
	  \bottomrule
	\end{tabular}
	\label{tab:VDEC}
\end{table}
\section{Conclusions}\label{sec:conc}
We introduced \ourmethodname, an open video generation system for automotive driving scenes. 
Our method is build entirely on open-source pre-trained general-purpose models, which we fine-tune to the automotive domain using the public driving videos dataset BDD100K. 
\ourmethodname{} employs dedicated models for image tokenization, world modeling, and video decoding being able to generate coherent 256×256-sized videos frame-by-frame at 4 fps with a single frame algorithmic latency (lookahead).
We motivate design choices by quantitative and qualitative analysis, and
 publish our training and inference code, thereby allowing for full reproducibility and deployment at academic scale.
\paragraph*{Acknowledgment} {\raggedright Computational resources were provided by the German AI Service Center \mbox{WestAI}.\par}
{
    \small
    \bibliographystyle{ieeenat_fullname}
    \bibliography{refs,manual_refs}
}
\clearpage

\newcommand{\imgpathBase}{figures/supplement/Supplement/collected/}
\newcommand{\imgpathBaseTwo}{figures/supplement/Supplement/collected2/}
\newcommand{\makesupplementtitle}{
    \twocolumn[{
    \begin{center}
    {\Large\bfseries  Supplementary Material: \\
    \ourmethodname: Video Generation for Automotive Driving Scenes\\ by Streamlining and Fine-Tuning Open Source Models with Public Data} \\[3ex]
    {\large
    Björn Möller \quad Zhengyang Li  \quad Malte Stelzer \quad Thomas Graave\\ Fabian Bettels \quad Muaaz Ataya \quad Tim Fingscheidt \\[0.3em]
    Technische Universität Braunschweig,
    Institute for Communications Technology}\\[0.3em]
    {\tt\small \{bjoern.moeller, zhengyang.li, malte.stelzer, thomas.graave,}\\
    {\tt\small f.bettels, m.ataya, t.fingscheidt\}@tu-bs.de}
    {\Large }
    \end{center}
            }]
}

\clearpage
\onecolumn
\makesupplementtitle


\renewcommand\thesection{\Alph{section}}
\setcounter{section}{0}  

\section{Details on used Open Source Models}
\paragraph{VQGAN} The vector-quantized generative adversarial network (\texttt{VQGAN}) combines the strengths of vector quantization \cite{van_den_oord_neural_2017}, autoencoders \cite{kingma_auto-encoding_2014}, and GANs \cite{goodfellow_generative_2014} to generate high-quality images with compact latent representations \cite{esser_taming_2021}.
The architecture consists of an encoder, a vector quantization module, and decoder, utilizing convolutional layers for image compression and reconstruction.
Without skip connections between encoder and decoder, both can be applied independently.
The open-source \texttt{VQGAN} model, which we leverage to initialize our fine-tuning process, has a spatial compression factor of 16, encoding to a latent dimension of $d\!=\!64$.
The vector quantizer (VQ) relies on nearest-neighbor lookup, using a vocabulary size $K \! = \! 8192 \! = \! 2^{13}$. 
The blocks ENC, VQ (jointly being TOK), and DEC utilize 58.7M, 532.4k, and 87M model parameters, respectively. 
Consequently, our TOK comprises 59.2M parameters, only about 20\% of the 300M parameters in GAIA-1's \cite{hu_gaia-1_2023} tokenizer.
The weights were originally optimized for 2.5M steps on real-world images of 256x256 resolution from various domains, yielding a general-purpose \texttt{VQGAN} \cite{patil_amused_2024}.
\textit{However, deploying it for front-facing camera images of automotive driving scenes, the model exhibits limited reconstruction quality.}

\paragraph{LWM}
The large world model (\texttt{LWM}) is a multimodal autoregressive decoder-only transformer capable of processing sequences as long as 1M tokens of both video and text modalities \cite{liu_world_2024}.
\texttt{LWM} is based on \texttt{LLaMA-2}  \cite{touvron_llama_2023} in its 7B parameter variant. 
Accordingly, it employs $N^{\mathrm{LWM}}\!=\!32$ decoder blocks and 32 heads in the multi-head attention. The internal feature dimension $\delta$ is 4,096.
The open-source \texttt{LWM-Chat-1M} \cite{liu_world_2024} model variant is used to initialize our WM. 
The authors initialized from \texttt{LLaMA-2} \cite{touvron_llama_2023} and trained on language tasks with text sequences of progressively increasing lengths up to 1M tokens.
In a second stage, they trained the \texttt{LWM} on four equally weighted vision-language tasks: image understanding, video understanding, text-image generation, and text-video generation.
The authors tokenized images and videos using the general-purpose \texttt{VQGAN} ENC+VQ \cite{patil_amused_2024}, also discussed above. 
Approximately 495B language-vision tokens were utilized, aggregated from datasets such as LAION-2B \cite{schuhmann_laion-5b_2022} and WebVid-10M \cite{bain_frozen_2021}, primarily consisting of captioned images and videos from diverse domains obtained via web scraping.
{\it However, no dedicated driving video dataset containing recordings from forward-facing cameras is utilized.
Also, the task of predicting future latent frames based on initial latent frames was not addressed in \texttt{LWM} training.
Moreover, \texttt{LWM} was trained with 256 tokens per image defining the image tokenizer output size and thus limits the video frame resolution, which we streamline to fit to the BDD100K automotive dataset.}

\section{Fine-Tuning the World Model}\label{B}
\begin{table*}[t]
	\centering
	\caption{\textbf{Tokenizer} and {\bf image decoder discriminator and loss weight} ablations: Quantitative results for using our $\mathrm{D}(\:)$ with varying weights $\lambda^{\prime}$, $\lambda^\mathrm{G}$ for perceptual loss and generator loss during fine-tuning with the total loss $J^{\mathrm{total}}$, employing $J^{D}$ for discriminator training, with $J^{\mathrm{SSL}}$ using \texttt{DINOv1} \cite{caron_emerging_2021}, evaluated on $\mathcal{D}_{\mathrm{BDDimg}}^{\mathrm{val}}$.  \texttt{VQGAN} models (TOK + DEC) were fine-tuned on  $\mathcal{D}_{\mathrm{BDDimg/5s}}^{\mathrm{train}}$ for 200k iterations.  Best results are in bold font, second best underlined. PSNR in (dB).
 }
	\begin{tabular}{lcc|c|cccccc}
		\toprule
        Discriminator  & $\lambda^{\prime}$ &    $\lambda^{G} \!$&  $\frac{\lambda^{\mathrm{G}}}{\lambda^{\prime}}$  & PSNR $\uparrow$	 & SSIM $\uparrow$ & MS-SSIM $\uparrow$ & LPIPS $\downarrow$ & FID  $\downarrow$ & CMMD $\downarrow$ \\
        \midrule
        $\mathrm{D}(\:)$ from \cite{esser_taming_2021} & 0.1 & 1.0  & 10 & 25.75 & 0.7630 & 0.9022  & 0.1170 & 5.48 & 0.074  \\
        \midrule
       Our $\mathrm{D}(\:)$ (Sec.\ 3 main paper) & 0.1 & 1.0  & 10 & \underline{25.88} & 0.7685  & 0.9039 & 0.1148 & 5.14 & 0.065  \\
       & 0.1 & 1.5   & 15 & \textbf{25.89} & 0.7694 & 0.9044 & 0.1146 & 5.32 & 0.064 \\
        & 0.1 & 3.0  & 30  & 25.85 & 0.7677 & 0.9041 & 0.1166  &5.34    & 0.075 \\
      \hspace{4em}. & 0.1 & 0.8  & 8 & 25.86 & 0.7682 & 0.9037 & 0.1146 & 5.02 & 0.064 \\
     & 0.3 & 1.0  & 3.3 & 25.68       & \underline{0.7704}    & \underline{0.9049}   & 0.1050 & 4.43 & 0.053 \\
      \hspace{4em}.  & 0.3 & 1.2 & 4 & 25.68 & \textbf{0.7712} & \textbf{0.9050} & 0.1053 & 4.45 & 0.059 \\
        \hspace{4em}. & 0.3 & 1.5  & 5 & 25.65 & 0.7696 & 0.9047 & 0.1055 & 4.50 & 0.057 \\
      & 0.5  & 1.0  & 2 & 25.47 & 0.7678 & 0.9030 & 0.1042 & 4.19 & 0.056 \\
       \hspace{4em} (\textbf{proposed}) & 1.0 & 1.0 & 1 & 25.14 & 0.7649 & 0.8991 & \textbf{0.1035} & 3.97 & \textbf{0.048} \\ 
    & 2.0 & 1.0 & 0.5 & 24.74 & 0.7616 & 0.8941 & 0.1042 & \textbf{3.55} & \underline{0.050} \\
    & 2.0 & 2.0 & 1 &  24.77 & 0.7623 & 0.8954 & \underline{0.1040} & \underline{3.93} & 0.051 \\
      \bottomrule
	\end{tabular}
	\label{tab:supplement_vqgan_discriminator_ablations}
\end{table*}
\paragraph{Fine-Tuning the World Model}
We fine-tune the world model (WM) on a 4 fps \textit{video} training dataset $\mathcal{D}_{\mathrm{BDDvid-4fps}}^{\mathrm{train-70k}}$ (cf.\ Table \ref{tab:dataset_bdd100k} in the main paper), as shown in Fig.\ 4 in the main paper for a single training sample.
The sample consists of a text token index sequence $c_1^M\!=\!(c_m)$ and an image patch token index sequence $k_1^{ (T+ N) \cdot n'} \!=\!(k_{\nu})$, which corresponds to $T\!=\!2$ initial video frames and $N\!=\!16$ future video frames, each represented by $n'\!=\!257$ indices.
Each index $k_{\nu} \in \mathcal{K}$, with the set of indices $\mathcal{K}\!=\!\{0,1,...,K\}$, represents either an entry from the image tokenizer's codebook $\mathrm{CB}$, or an end-of-frame token ($k_{\nu}\!=\!0$).
As model input, the text index sequence $c_1^M$ and the image token index sequence $k_1^{(T+ N) \cdot n'-1} \!=\!(k_{\nu})$ that omits the last ground-truth index are both embedded with dedicated layers into a $\delta\!=\!4096$-dimensional space before being concatenated.
The resulting input sequence $\mathbf{y}_1^L\!=\!(\mathbf{y}_\ell)$ of length $L \! = \! M + (T + N ) \cdot n' -1$ is then processed by the WM's $N^{\mathrm{LWM}}$ decoder blocks.
To fine-tune for next-token prediction, decoder blocks apply causal attention to mask future tokens, forcing the WM to rely solely on preceding and current elements of the (ground-truth) input sequence. 
This teacher-forcing mechanism differs from autoregressive inference, which relies on prior index predictions.
Next, a root mean square layer normalization (RMS-Norm) \cite{zhang_root_2019} and an $\mathrm{FC(}K+1\mathrm{)}$ layer with softmax activation are applied to each position in the sequence. 
The output sequence $\mathbf{P}_{2}^{L+1}$ comprises $L$ discrete probability distributions $\mathbf{P}_{\ell} \in [0,1]^{K+1}$, each estimating the token probabilities at ground-truth sequence position $\ell \in \mathcal{L}\!=\!\{2,...,L\!+\!1\}$.
For loss calculation, we cut this sequence to contain only the predicted probability distributions $\mathbf{P}_{M+1}^{L+1}\!=\!(\mathbf{P}_\ell)$ for \textit{image patch} token indices corresponding to the ground-truth image patch token indices $k_1^{(T +N) \cdot n'} \!=\!k_{\nu}$, and optimize using their respective cross-entropy loss
\begin{equation}\label{eq:loss_Jwm}
J^{\mathrm{CE}} = - \frac{1}{(T + N) \cdot n^{\prime}} \sum_{\nu\in\mathcal{I}} \mathrm{log} P(k_\nu \vert c_1^{M},k_1^{\nu-1}),
\end{equation}
where $\mathcal{I}\!=\!\{1,2,...,(T+N)\cdot n'\}$ is used to iterate over all image token indices in the target sequence and the estimated probability for the ground-truth token index $k_\nu$ corresponds to $P(k_\nu \vert c_1^{M}, k_1^{\nu-1})$.

Due to the size of the WM, full fine-tuning is impractical, so we use the parameter-efficient LoRA \cite{hu_lora_2022} method, fine-tuning only a small set of additional adaptor weights for all linear and embedding layers, with full fine-tuning limited to normalization layer weights.
This results in 6.97B total model parameters, with only 2.39\% of these being fine-tuneable.
To further reduce memory usage, frozen parameters are stored in the lower-precision format \texttt{bfloat16}, while only fine-tunable parameters remain in \texttt{FP32}.

\section{Fine-Tuning the Tokenizer and Image Decoder}\label{sec:C}

\begin{figure*}[t]
    \centering
    \setlength{\tabcolsep}{2pt} 
    \renewcommand{\arraystretch}{1.2} 
    \begin{tabular}{c c c c c c}
        \raisebox{5\height}{(a) Original} &
        \includegraphics[width=0.16\textwidth]{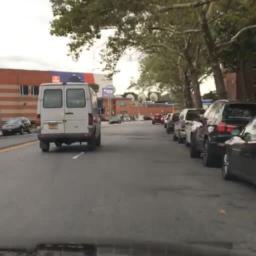} &
        \includegraphics[width=0.16\textwidth]{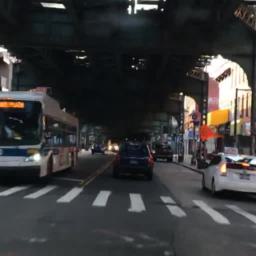} &
        \includegraphics[width=0.16\textwidth]{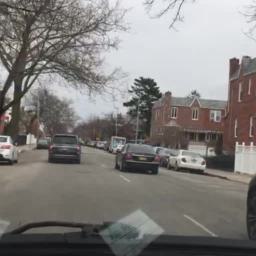} &
        \includegraphics[width=0.16\textwidth]{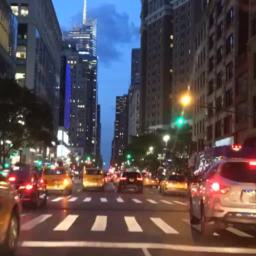} &
        \includegraphics[width=0.16\textwidth]{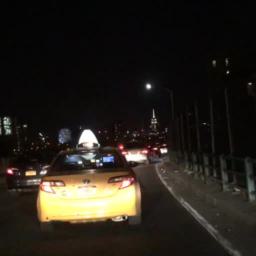}\\
        
        \raisebox{5\height}{(b) No Fine-tuning} &
        \includegraphics[width=0.16\textwidth]{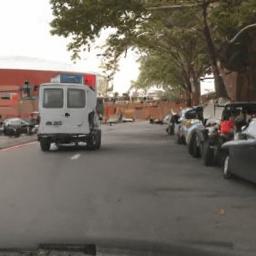} &
        \includegraphics[width=0.16\textwidth]{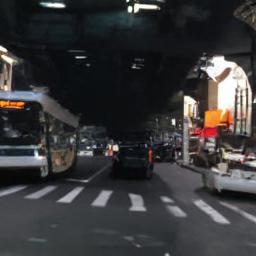} &
        \includegraphics[width=0.16\textwidth]{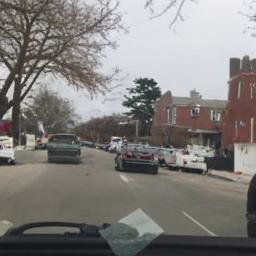} &
        \includegraphics[width=0.16\textwidth]{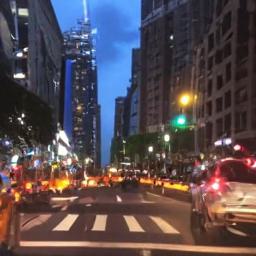} &
        \includegraphics[width=0.16\textwidth]{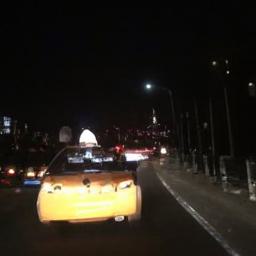} \\

        \raisebox{5\height}{{(c) $J^{\mathrm{total}}$}} &
        \includegraphics[width=0.16\textwidth]{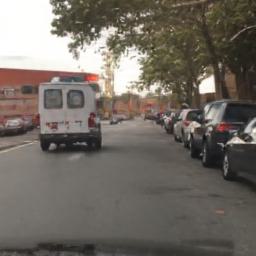} &
        \includegraphics[width=0.16\textwidth]{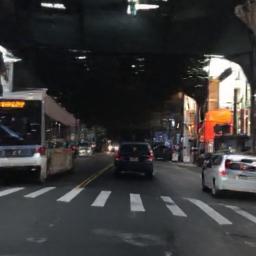} &
        \includegraphics[width=0.16\textwidth]{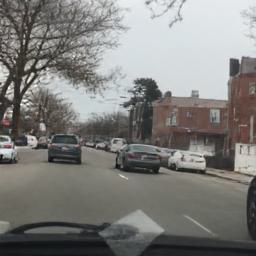} &
        \includegraphics[width=0.16\textwidth]{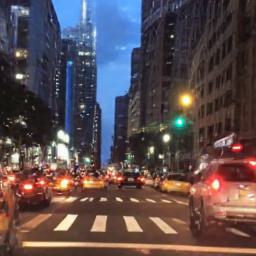} &
        \includegraphics[width=0.16\textwidth]{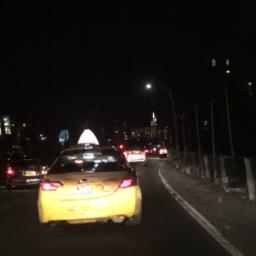} \\

        \raisebox{5\height}{(d) ... -$J^{\mathrm{SSL}}$} &
        \includegraphics[width=0.16\textwidth]{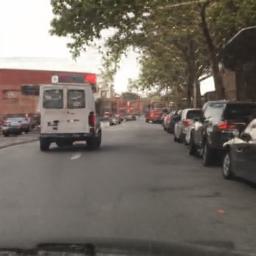} &
        \includegraphics[width=0.16\textwidth]{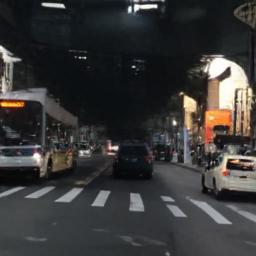} &
        \includegraphics[width=0.16\textwidth]{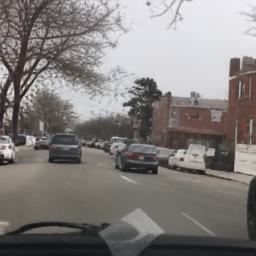} &
        \includegraphics[width=0.16\textwidth]{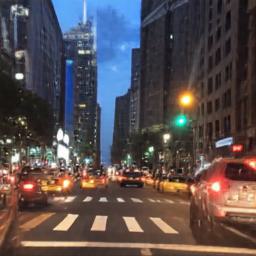} &
        \includegraphics[width=0.16\textwidth]{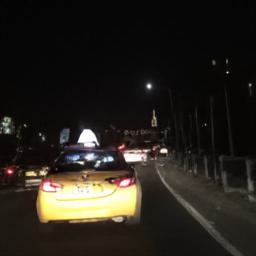} \\

        \raisebox{5\height}{(e) ... -$J^{\mathrm{\prime}}$} &
        \includegraphics[width=0.16\textwidth]{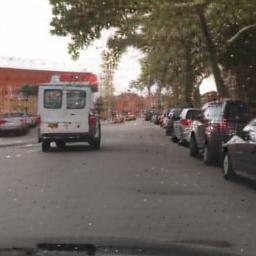} &
        \includegraphics[width=0.16\textwidth]{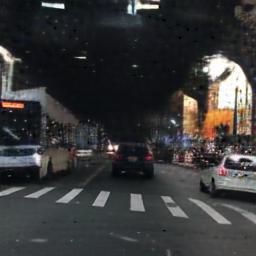} &
        \includegraphics[width=0.16\textwidth]{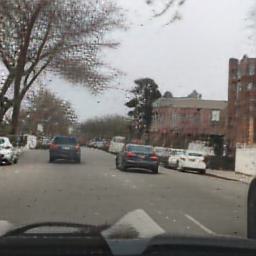} &
        \includegraphics[width=0.16\textwidth]{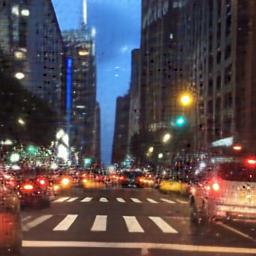} &
        \includegraphics[width=0.16\textwidth]{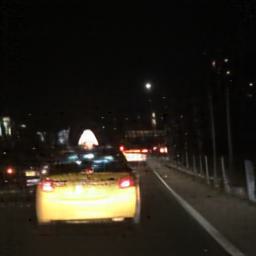} \\

        \raisebox{5\height}{(f) ... -$J^{\mathrm{L2}}$} &
        \includegraphics[width=0.16\textwidth]{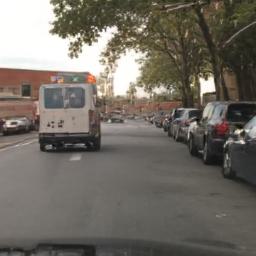} &
        \includegraphics[width=0.16\textwidth]{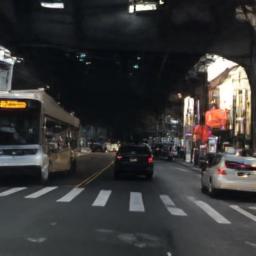} &
        \includegraphics[width=0.16\textwidth]{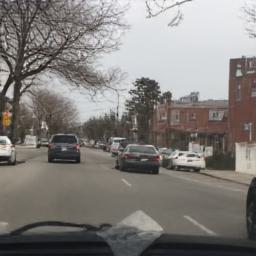} &
        \includegraphics[width=0.16\textwidth]{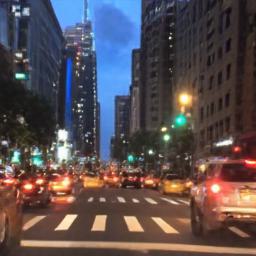} &
        \includegraphics[width=0.16\textwidth]{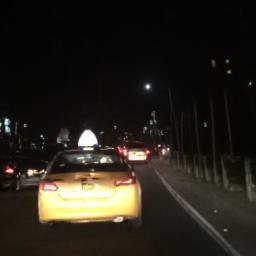} \\

        \raisebox{5\height}{(g) ... -$J^{\mathrm{G}}$} &
        \includegraphics[width=0.16\textwidth]{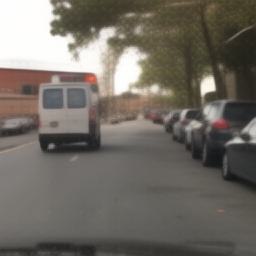} &
        \includegraphics[width=0.16\textwidth]{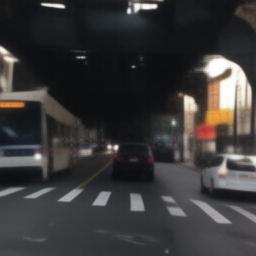} &
        \includegraphics[width=0.16\textwidth]{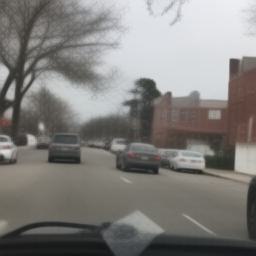} &
        \includegraphics[width=0.16\textwidth]{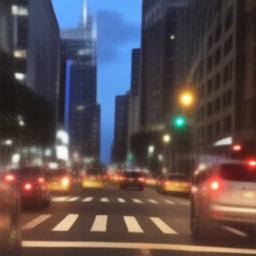} &
        \includegraphics[width=0.16\textwidth]{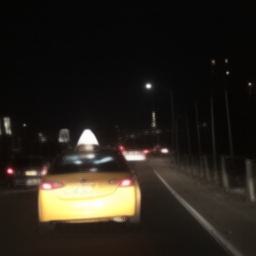} \\
        
    \end{tabular}
   \caption{\textbf{Visual examples} of transcoded (TOK+DEC) validation images for a (b) pre-trained \texttt{VQGAN} and various fine-tuned versions employing (c) $J^{\mathrm{total}}$, and (d-g) $J^{\mathrm{total}}$ with individually omitted (-$J$) loss components corresponding to Table \ref{tab:vqgan_loss_ablations} in the main paper. In the first row (a) original images are from $\mathcal{D}_{\mathrm{BDDimg}}^{\mathrm{val}}$. Best viewed digitally with zoom.
}
\label{fig:supp_tokenizer_quali}
\end{figure*}

Fig.\ \ref{fig:supp_tokenizer_quali} shows visual examples of an original validation image (Fig.\ \ref{fig:supp_tokenizer_quali}a), the transcoded (TOK+DEC) output using the pre-trained, not fine-tuned \texttt{VQGAN} (Fig.\ \ref{fig:supp_tokenizer_quali}b) and various reconstructions for fine-tuned models corresponding to Table \ref{tab:vqgan_loss_ablations} from the main paper.
We find the results of the fine-tuning with $J^{\mathrm{total}}$ most visually convincing in terms of perceptual image quality. which is in line with the quantitative evaluations. 
Omitting the self-supervised learning component $J^{\mathrm{SSL}}$ (Fig. \ref{fig:supp_tokenizer_quali}d) appears to also produce good results, although the image quality is slightly less refined.
Excluding the perceptual loss term $J^{\mathrm{\prime}}$ (Fig.\ \ref{fig:supp_tokenizer_quali}e) results in a noticeable degradation in perceptual quality, with visible reconstruction artifacts.
Removing the $J^{\mathrm{L2}}$ loss term (Fig.\ \ref{fig:supp_tokenizer_quali}f) leads to sharp but excessively detailed image regions compared to $J^{\mathrm{total}}$.
Finally, omitting the generator component of the GAN loss $J^{\mathrm{G}}$ (Fig.\ \ref{fig:supp_tokenizer_quali}g)clearly causes blurred reconstructions that lack fine details.
Table \ref{tab:supplement_vqgan_discriminator_ablations} extends Table \ref{tab:vqgan_discriminator_ablations} from the main paper. We show more experiments on loss weights $\lambda^{\prime}$ and $\lambda^{\mathrm{G}}$ of the important $J^{\mathrm{total}}$ contributions $J^{\prime}$ and $J^{\mathrm{G}}$, respectively.

\newcommand{\ImgColumnTwo}{b1d0a191-06deb55d.jpg.jpg}

\section{Generated Video Examples}\label{sec:D}
\begin{figure*}[ht]
    \centering
    \hspace{10mm}
    \setlength{\unitlength}{1mm}
    \begin{overpic}[width=0.9\linewidth,grid=false]{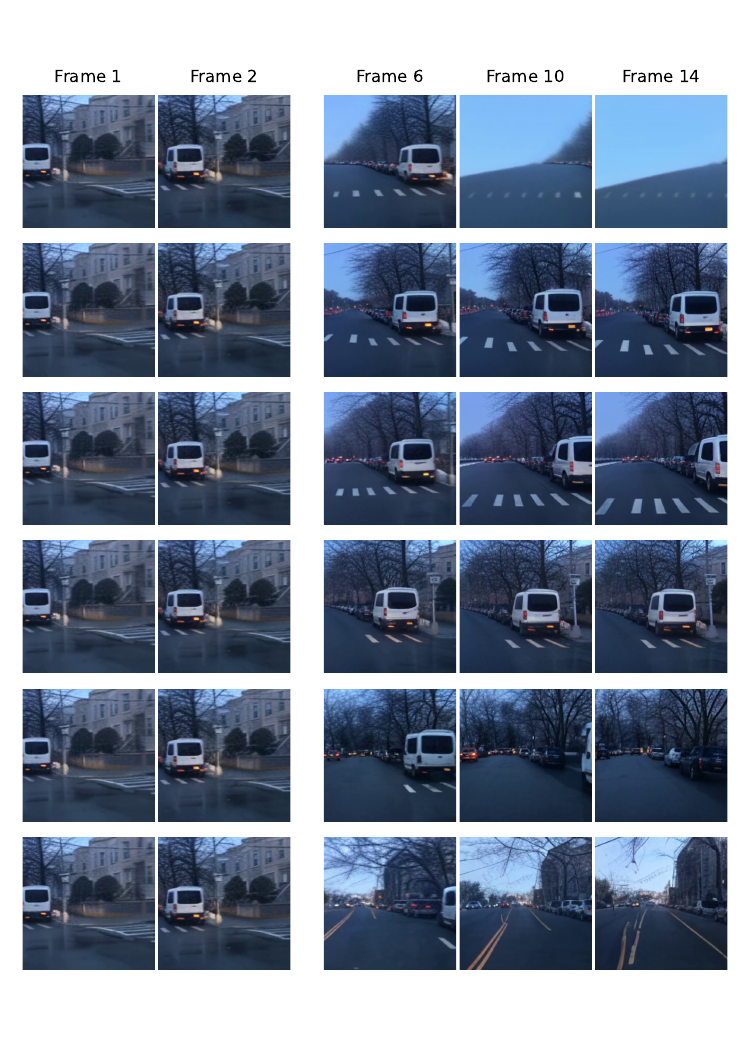}
        \hspace{80mm}
        \put(2,94){\color{gray!60}\rule{55mm}{1.5mm}} 
         \put(31,94){\color{gray!60}\rule{84mm}{1.5mm}}
        \put(3,96){Input Frames}
        \put(32,96){Generated Frames}
        
        
        \put(-7.5,84){top-$k$=1}
        \put(-7.5,70){top-$k$=5}
        \put(-7.5,55){top-$k$=10}
        \put(-7.5,42){top-$k$=50}
        \put(-7.5,28){top-$k$=200}
        \put(-7.5,13){top-$k$=1000}
    \end{overpic}
     \vspace{-12mm} 
    \caption{\textbf{Generated videos} using \textbf{OpenViGA}: Each row shows a video generated with a different top-$k$ sampling parameter, where $k \in \{1,5,10,50,200,1000\}$. The first $T\!=\!2$ frames in each row represent the input frames from $\mathcal{D}_{\mathrm{BDDvid-4fps}}^{\mathrm{val-500}}$, while the other three frames represent the world model’s prediction after 1 (Frame 6), 2 (Frame 10), and 3 (Frame 14) seconds into the future, respectively. Note that input frames are transcoded using our TOK+DEC. Best viewed digitally with zoom.}
    \label{fig:topk_example1}
\end{figure*}

\begin{figure*}[ht]
    \centering
    \hspace{10mm}
    \setlength{\unitlength}{1mm}
    \begin{overpic}[width=0.9\linewidth,grid=false]{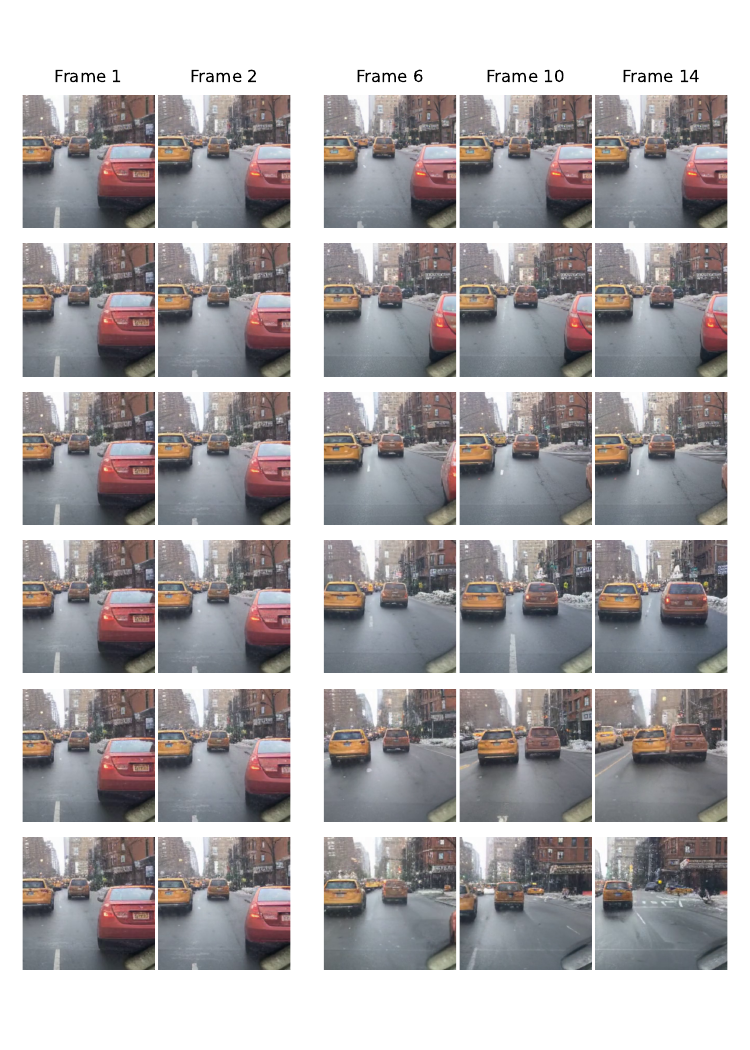}
        \hspace{80mm}
        \put(2,94){\color{gray!60}\rule{55mm}{1.5mm}} 
         \put(31,94){\color{gray!60}\rule{84mm}{1.5mm}}
        \put(3,96){Input Frames}
        \put(32,96){Generated Frames}
        
        
        \put(-7.5,85){top-$k$=1}
        \put(-7.5,70){top-$k$=5}
        \put(-7.5,55){top-$k$=10}
        \put(-7.5,42){top-$k$=50}
        \put(-7.5,28){top-$k$=200}
        \put(-7.5,13){top-$k$=1000}
    \end{overpic}
     \vspace{-12mm} 
    \caption{\textbf{Generated videos} using \textbf{OpenViGA}: Each row shows a video generated with a different top-$k$ sampling parameter, where $k \in \{1,5,10,50,200,1000\}$. The first $T\!=\!2$ frames in each row represent the input frames from $\mathcal{D}_{\mathrm{BDDvid-4fps}}^{\mathrm{val-500}}$, while the other three frames represent the world model’s prediction after 1 (Frame 6), 2 (Frame 10), and 3 (Frame 14) seconds into the future, respectively. Note that input frames are transcoded using our TOK+DEC. Best viewed digitally with zoom.}
    \label{fig:topk_example2}
\end{figure*}

Fig.\ \ref{fig:topk_example1} and Fig.\ \ref{fig:topk_example2} show generated videos using OpenViGA, each generated with a different top-$k$ parameter, corresponding to Table \ref{tab:lwm_ablations} in the main paper.


\clearpage

\end{document}